\documentclass{article}

\usepackage{arxiv}

\usepackage[utf8]{inputenc} 
\usepackage[T1]{fontenc}    
\usepackage{hyperref}       
\usepackage{url}            
\usepackage{booktabs}       
\usepackage{amsfonts}       
\usepackage{nicefrac}       
\usepackage{microtype}      
\usepackage{lipsum}
\usepackage{graphicx}
\graphicspath{ {./images/} }

\usepackage{amsthm}
\usepackage{amsmath}
\usepackage{amsfonts}
\usepackage{amssymb}
\usepackage{xspace}
\usepackage{verbatim}
\usepackage{xcolor}
\usepackage{bbold}
\usepackage{subcaption}
\usepackage{float}
\usepackage{tikz}
\usepackage{hyperref}
\usepackage{longtable}
\usepackage{multirow}
\usepackage{stackengine}
\usepackage{pdflscape}
\usepackage{graphicx}
\usepackage{bm}
\usepackage{mathrsfs}
\usepackage{stmaryrd}
\usepackage{pdflscape}

\newtheorem{remark}{Remark}[section]

\definecolor{revisar}{rgb}{0.19, 0.55, 0.91}

\newcommand\Tstrut{\rule{0pt}{2.6ex}}         
\mathchardef\mhyphen="2D
\newcommand{\Ione}{{\bf(I1)}\xspace}
\newcommand{\Itwo}{{\bf(I2)}\xspace}
\newcommand{\Ithree}{{\bf(I3)}\xspace}

\newcommand{\DF}{{\bf(DF)}\xspace}

\newcommand{\DC}{{\bf(DC)}\xspace}
\newcommand{\SBC}{{\bf(SBC)}\xspace}
\newcommand{\IP}{{\bf(IP)}\xspace}
\newcommand{\OP}{{\bf(OP)}\xspace}
\newcommand{\LOP}{{\bf(LOP)}\xspace}
\newcommand{\ROP}{{\bf(ROP)}\xspace}
\newcommand{\OPe}{{\bf(OP$_{\bm{e}}$)}\xspace}
\newcommand{\NP}{{\bf(NP)}\xspace}
\newcommand{\NPe}{{\bf(NP$_{\bm{e}}$)}\xspace}
\newcommand{\EP}{{\bf(EP)}\xspace}
\newcommand{\PEP}{{\bf(PEP)}\xspace}
\newcommand{\EPOne}{{\bf(EP1)}\xspace}
\newcommand{\IB}{{\bf(IB)}\xspace}
\newcommand{\SIB}{{\bf(SIB)}\xspace}
\newcommand{\CB}{{\bf(CB)}\xspace}
\newcommand{\LF}{{\bf(LF)}\xspace}
\newcommand{\LT}{{\bf(LT)}\xspace}
\newcommand{\SPP}{{\bf(SP)}\xspace}
\newcommand{\ISPP}{{\bf(ISP)}\xspace}
\newcommand{\MI}{{\bf(MI)}\xspace}
\newcommand{\IFI}{{\bf(IFI)}\xspace}
\newcommand{\TT}{{\bf(TT)}\xspace}
\newcommand{\C}{{\bf(C)}\xspace}
\newcommand{\CPN}{{\bf(CP(N))}\xspace}
\newcommand{\LCPN}{{\bf(L$\mhyphen$CP(N))}\xspace}
\newcommand{\RCPN}{{\bf(R$\mhyphen$CP(N))}\xspace}
\newcommand{\LI}{{\bf(LI$_{\bm{T}}$)}\xspace}
\newcommand{\WLI}{{\bf(WLI$_{\bm{F}}$)}\xspace}
\newcommand{\GLI}{{\bf(GLI)}\xspace}
\newcommand{\CLI}{{\bf(CLI)}\xspace}
\newcommand{\TNMT}{{\bf(MT$_{\bm{T},\bm{N}}$)}\xspace}
\newcommand{\RP}{\bf(RP)\xspace}
\newcommand{\GEP}{\bf(GEP)\xspace}
\newcommand{\ME}{\bf(ME)\xspace}
\newcommand{\GHS}{{\bf(GHS)}\xspace}
\newcommand{\TC}{{\bf(TC)}\xspace}
\newcommand{\OC}{{\bf(OC)}\xspace}
\newcommand{\UC}{{\bf(UC)}\xspace}
\newcommand{\PIT}{{\bf(PI$_{\bm{T}}$)}\xspace}
\newcommand{\PIIT}{{\bf(PII$_{\bm{T}}$)}\xspace}
\newcommand{\FOP}{{\bf(FOP)}\xspace}

\newcommand{\IPe}{{\bf(IP$_{\bm{e}}$)}\xspace}

\newcommand{\DST}{{\bf(D$\mhyphen$ST)}\xspace}
\newcommand{\DTS}{{\bf(D$\mhyphen$TS)}\xspace}
\newcommand{\DTT}{{\bf(D$\mhyphen$TT)}\xspace}
\newcommand{\DSS}{{\bf(D$\mhyphen$SS)}\xspace}

\newcommand{\NDOne}{\ensuremath{N_{\bm{D_1}}}\xspace}

\newcommand{\IYG}{\ensuremath{I_{\bm{YG}}}\xspace}
\newcommand{\IRC}{\ensuremath{I_{\bm{RC}}}\xspace}
\newcommand{\IRS}{\ensuremath{I_{\bm{RS}}}\xspace}
\newcommand{\ILK}{\ensuremath{I_{\bm{LK}}}\xspace}

\newcommand{\IGD}{\ensuremath{I_{\bm{GD}}}\xspace}
\newcommand{\IGG}{\ensuremath{I_{\bm{GG}}}\xspace}
\newcommand{\IKD}{\ensuremath{I_{\bm{KD}}}\xspace}
\newcommand{\IWB}{\ensuremath{I_{\bm{WB}}}\xspace}
\newcommand{\IFD}{\ensuremath{I_{\bm{FD}}}\xspace}
\newcommand{\ILT}{\ensuremath{I_{\bm{Lt}}}\xspace}
\newcommand{\IGT}{\ensuremath{I_{\bm{Gt}}}\xspace}

\newcommand{\xt}[2]{{#1}_{T}^{(#2)}}

\usepackage{pifont}
\usepackage{tabularx}
\def\mystrut(#1,#2){\vrule height #1 depth #2 width 0pt}
\newcolumntype{C}[1]{%
	>{\mystrut(3ex,2ex)\centering}%
	p{#1}%
	<{}}

\title{A Comprehensive Survey of Fuzzy Implication Functions}

\author{
 Raquel Fernandez-Peralta \\
  Mathematical Institute\\
  Slovak Academy of Sciences\\
  Bratislava, Slovakia \\
  \texttt{raquel.fernandez@mat.savba.sk}
}

\begin{document}
\maketitle
\begin{abstract}
Fuzzy implication functions are a key area of study in fuzzy logic, extending the classical logical conditional to handle truth degrees in the interval $[0,1]$. While existing literature often focuses on a limited number of families, in the last ten years many new families have been introduced, each defined by specific construction methods and having different key properties. This survey aims to provide a comprehensive and structured overview of the diverse families of fuzzy implication functions, emphasizing their motivations, properties, and potential applications. By organizing the information schematically, this document serves as a valuable resource for both theoretical researchers seeking to avoid redundancy and practitioners looking to select appropriate operators for specific applications.
\end{abstract}


\newpage

\tableofcontents 

\newpage

\section{Introduction}\label{sec:introduction}

One of the most important branches of fuzzy logic corresponds to the study of fuzzy operators, which are used to operate between membership values or truth degrees. Traditionally, many fuzzy concepts were defined as a generalization of the corresponding one in classical logic. Following this reasoning, the main classical logic connectives have been generalized: the intersection or conjunction is defined as a fuzzy conjunction (usually a t-norm); the union or disjunction is defined as a fuzzy disjunction (usually a t-conorm); the negation or the complement is defined as a fuzzy negation; and the conditionals are represented by fuzzy implication functions. However, the study of fuzzy operators goes beyond logic connectives and it intersects with the study of aggregation functions. Aggregation functions (also called aggregation operators) are used for combining and merging values into a single one according to a certain objective. Since fuzzy operators play an important role in a wide variety of applications, many different types have been defined. To illustrate this fact we refer the reader to some books exclusively devoted to this topic \cite{Klement2000,Calvo2002,Baczynski2008,Beliakov2010,Grabisch2009,Alsina2006}. Although other domains besides $[0,1]$ have been considered in the literature \cite{Goguen1967,Munar2023}, typically fuzzy operators are defined as functions $F:[0,1]^n \to [0,1]$ that fulfill some set of conditions (monotonicity, continuity, associativity, commutativity, boundary conditions...). However, these conditions are usually general enough to allow the existence of many different operators of a certain kind. This results in the more specific study of different classes of operators that fulfill a certain set of conditions, in which desired additional properties apart from the ones in the operator's definition can be included.

Fuzzy implication functions are defined as functions $I:[0,1]^2 \to [0,1]$ which are decreasing with respect to the first variable, increasing with respect to the second variable and they coincide with the classical implication in $\{0,1\}^2$ \cite{Baczynski2008,Fodor1994}. In the same way boolean implications are employed in inference schemas like modus ponens, modus tollens, etc., fuzzy implication functions play a similar role in the generalization of these schemas modeling the corresponding conditionals which are called fuzzy IF-THEN rules. These rules are widely used in approximate reasoning, wherein from imprecise inputs and fuzzy premises or rules, imprecise conclusions are drawn. However, apart from inference systems based on fuzzy rules \cite{Combs1998,Jayaram2008,Jayaram2008B}, fuzzy implication functions are also considered in other application areas like fuzzy mathematical morphology or data mining \cite{Baczynski2015}. 

Partly motivated by their potential applications, the study of fuzzy implication functions has significantly grown in the last decades (see the bibliometric analysis in \cite{Laenge2021}). Indeed, some monographs \cite{Baczynski2008,Baczynski2013} and surveys \cite{Baczynski2015,Mas2007,Baczynski2008B} only devoted to the study of these operators have been published. However, all these studies have one thing in common, they center their efforts to only a few families of fuzzy implication functions. If one takes a quick glance to the existing bibliography on this topic one can quickly realize that nowadays much more families have been defined apart from the ones considered in the existing monographs.

One of the main justifications when introducing a new family of fuzzy implication functions is that depending on the context and the concrete applications, different properties of the operator are needed, so it is important to have different options from where to choose when using fuzzy implication functions in applications. In this sense, the situation presented above could appear very appealing, since nowadays we have much more families among to choose than years before. However, it has been discussed several times that although a ``new" family presents a novel construction method, it might have intersection with other families on the literature or can even completely coincide with an already existing family. For this reason, avoiding redundancy is an important aspect to take into account when defining a new family of these operators. Some couple of well-practices are commented by the experts in order to avoid redundancy \cite{Massanet2024}, the more relevant are the study of characterizations and the intersections with other families.

Thus, this survey has a different perspective from the existing ones, our main objective is to collect as many families and additional properties as possible and to highlight the motivation behind its introduction.
 This document pretends to be a good consulting document, with a schematic structure, in order to know the different families available. This document can be useful for two perspectives: for the theoretical perspective that studies the families of fuzzy implication functions to avoid redundancy, to consult what has been already done, and for the application perspective,  to easily see the possible operators and to choose among them for possible practical applications.

 The structure of the paper is as follows: in Section \ref{sec:additional_properties} we list the additional properties of fuzzy implication functions, in Section \ref{sec:families} we gather the families of fuzzy implication function distinguishing between several classes and subclasses and the document ends in Section \ref{sec:conclusions} with some conclusions and future prospects.


\section{Additional properties}\label{sec:additional_properties}

Since the definition of fuzzy implication function is quite general, additional properties of these operators are usually considered. These properties come often in the form of functional equations which involve fuzzy implication functions and some of them, other operators as well. The motivation behind the definition of these additional properties are diverse, but the most usual are: a large majority of them were introduced as the straightforward generalizations of classical logic tautologies to fuzzy logic: others point out some desirable or interesting analytical/algebraic properties of these functions: some were introduced since they appeared when solving a particular problem: many are generalizations of other additional properties: finally, a lot of these properties aim to be useful in a particular problem or application. In Table \ref{table:additional_properties} the reader can find a list with several additional properties of fuzzy implication functions introduced throughout the years. As far as possible, we have included the motivation behind the introduction of the property and related studies. 


\begin{longtable}{|p{0.4\textwidth}|p{0.4\textwidth}|p{0.1\textwidth}|}
	\hline
	\textbf{Name and Expression} & \textbf{Comments and References} & \textbf{ID}  \Tstrut \\ \hline

	\textbf{First place antitonicity}
	$$I(x_1,y) \geq I(x_2,y),$$ 
        $x_1,x_2,y \in [0,1]$ such that $x_1 \leq x_2$. & States that a fuzzy implication function should be decreasing with respect to the first variable.  &  \Ione \Tstrut \\ \hline

	\textbf{Second place isotonicity}
	$$I(x,y_1) \leq I(x,y_2),$$ 
        $x,y_1,y_2 \in [0,1]$ such that $y_1 \leq y_2$. & States that a fuzzy implication function should be increasing with respect to the second variable.  &  \Itwo \Tstrut \\ \hline

	\textbf{Boundary conditions}
	$$I(0,0)=I(1,1)=1,\quad I(1,0)=0.$$ & States that a fuzzy implication function should satisfy the boundary conditions of the crisp implication. & \Ithree \Tstrut \\ \hline

	\textbf{Dominance of falsity of antecedent}
	$$I(0,y)=1, \quad y \in [0,1]$$ 
         &   &  \DF \Tstrut \\ \hline

	\textbf{Dominance of truth of consequent}
	$$I(x,1)=1, \quad x \in [0,1]$$ 
         &   &  \DC \Tstrut \\ \hline 

	\textbf{Strong boundary condition for $0$}
	$$I(x,0)=0, \quad x \in (0,1].$$
        Which is equivalent to impose $N_I = \NDOne$.
         & \cite{Dimuro2015}  &  \SBC \Tstrut \\ \hline  
	
	\textbf{Identity Principle}
	$$I(x,x)=1, \quad x \in [0,1]$$ & States that the overall truth value should be 1	when the truth values of the antecedent and the consequents are equal and can
	be seen as the generalization of the following tautology from the classical logic:
	$$ p \to p $$  &  \IP \Tstrut \\ \hline

	\textbf{Identity Principle with respect to $e \in [0,1]$}
	$$I(x,x) \geq e, \quad x \in [0,1]$$ & It Generalizes the property \IP for any $e \in [0,1]$. These kind of generalizations are usually related to fuzzy implication functions generated by some aggregation function with neutral element $e$, like $(U,N)$-implications or $U$-implications \cite{Li2015B,Li2015}. & \IPe
	\Tstrut \\ \hline
	
	\textbf{Ordering Property}
	$$I(x,y)=1 \Leftrightarrow x \leq y, \quad x,y\in[0,1]$$ & Imposes
	an ordering on the underlying set. & \OP
	\Tstrut \\ \hline

	\textbf{Left Ordering Property}
	$$I(x,y)=1 \Rightarrow x \leq y, \quad x,y\in[0,1]$$ & \cite{Zhou2020} & \LOP
	\Tstrut \\ \hline

	\textbf{Right Ordering Property}
	$$x \leq y \Rightarrow I(x,y)=1, \quad x,y\in[0,1]$$ & \cite{Zhou2020} & \ROP
	\Tstrut \\ \hline

	\textbf{Flexible ordering property with respect to $\theta$}
	$$x \leq \theta(y) \Rightarrow I(x,y)=1, \quad x,y\in[0,1],$$
    where $\theta:[0,1] \to [0,1]$ is a continuous and strictly increasing function with $\theta(1)=1$.& \cite{Zhou2021} & \FOP
	\Tstrut \\ \hline

	\textbf{Ordering Property with respect to $e \in [0,1]$}
	$$I(x,y) \geq e \Leftrightarrow x \leq y, \quad x,y\in[0,1]$$ & It Generalizes the property \OP for any $e \in [0,1]$. These kind of generalizations are usually related to fuzzy implication functions generated by some aggregation function with neutral element $e$, like $(U,N)$-implications or $U$-implications \cite{Li2015B,Li2015}. & \OPe
	\Tstrut \\ \hline
	
	\textbf{Left Neutrality Property}
	$$I(1,y)=y, \quad y\in[0,1]$$ & Captures the notion that a tautology allows the truth value of the consequent to be assigned as the overall truth value of the statement. Generalization of the classical tautology
	known as the exchange principle:
	$$ (1 \to p) \equiv p.$$ & \bf{(NP)}
	\Tstrut \\ \hline

	\textbf{Left Neutrality Property with respect to $e \in [0,1]$}
	$$I(e,y)=y, \quad y\in[0,1]$$ & It generalizes the property \NP for any $e \in [0,1]$. These kind of generalizations are usually related to fuzzy implication functions generated by some aggregation function with neutral element $e$, like $(U,N)$-implications or $U$-implications \cite{Li2015B,Li2015}. & \NPe
	\Tstrut \\ \hline
 
	\textbf{Exchange principle}
	$$I(x,I(y,z)) = I(y,I(x,z)), \quad  x,y,z\in[0,1]$$  & Generalization of the classical tautology
	known as the exchange principle:
	$$ p \to (q \to r ) \equiv q \to (p \to r).$$
        In \cite{Jayaram2011} the authors characterize the residuals that satisfy \EP.
        & \EP
	\Tstrut \\ \hline

	\textbf{Pseudo-exchange principle}
	$$I(x,z) \geq y \Leftrightarrow I(y,z) \geq x, \quad  x,y,z\in[0,1]$$  & \cite{Dimuro2015}
        & \PEP
	\Tstrut \\ \hline

	\textbf{Exchange principle for 1}
	$$I(x,I(y,z)) = 1 \Rightarrow I(y,I(x,z))=1,$$ $x,y,z\in[0,1].$  & \cite{Dimuro2015}
        & \EPOne
	\Tstrut \\ \hline

	\textbf{Generalized exchange property}
 
        Let $I,J$ be two fuzzy implication functions
        $$I(x,J(y,z)) = I(y,J(x,z)), \quad x,y,z \in [0,1]$$
        & 
        To generalize the exchange property to a pair of fuzzy implications \cite{Reiser2013}.
	& \GEP
	\Tstrut \\ \hline

	\textbf{Mutual exchangeability}
 
        Let $I,J$ be two fuzzy implication functions
        $$I(x,J(y,z)) = J(y,I(x,z)), \quad x,y,z \in [0,1]$$
        & 
        To generalize the exchange property to a pair of fuzzy implications \cite{Vemuri2015}.
	& \ME
	\Tstrut \\ \hline

	\textbf{Iterative Boolean Law} 
	$$ I(x,y)=I(x,I(x,y)), \quad x,y \in [0,1].$$&
	Generalization of the classical tautology:
	$$ p \to (p \to q) \equiv p \to q.$$ & \IB
	\Tstrut \\ \hline

	\textbf{Sub-iterative Boolean Law} 
	$$ I(x,y) \leq I(x,I(x,y)), \quad x,y \in [0,1].$$& \cite{Dimuro2015}
	 & \SIB
	\Tstrut \\ \hline
	
	\textbf{Consequent Boundary} 
	$$ I(x,y) \geq y, \quad x,y \in [0,1]$$&
	 & \CB
	\Tstrut \\ \hline
	
	\textbf{Lowest falsity} 
	$$ I(x,y)=0 \Leftrightarrow x=1 \text{ and } y=0.$$& It is useful when constructing strong equality indexes \cite{Bustince2013}.
	& \LF
	\Tstrut \\ \hline
	
	\textbf{Lowest truth} 
	$$ I(x,y)=1 \Leftrightarrow x=0 \text{ and } y=1.$$& It is useful when constructing strong equality indexes \cite{Bustince2013}. \vspace{2.5cm}
	& \LT \Tstrut \\ \hline

	\textbf{Specialty} 
	For any $\varepsilon > 0$ and for all $x,y \in [0,1]$ such that $x+\varepsilon,y+\varepsilon \in [0,1]$
 $$ I(x,y) \leq I(x+\varepsilon, y+\varepsilon).$$
& 
It imposes that the operator is monotonic increasing with respect both variables together \cite{Mis2017}. Also, this property is related to special GUHA-implicative quantifiers \cite{Sainio2008}. Also, the property is deeply studied in \cite{Jayaram2009}.
	& \SPP \Tstrut \\ \hline

	\textbf{Inverse specialty} 
        For any $\varepsilon > 0$ and for all $x,y \in [0,1]$ such that $x+\varepsilon,y+\varepsilon \in [0,1]$
	$$ I(x,y) \leq I(x+\varepsilon, y+\varepsilon).$$ &  It imposes that the operator is monotonic decreasing with respect both variables together \cite{Mis2017}.
	& \ISPP \Tstrut \\ \hline
 
	\textbf{$\alpha$-migrativity} 
	Let $\alpha \in (0,1)$ fixed.
	$$ I(x\alpha,y)=I(x,1-\alpha+\alpha y), \quad x,y \in [0,1].$$& To consider the well-known property of $\alpha$-migrativity studied for aggregation functions \cite{Bustince2012} in the case of fuzzy implication functions \cite{Baczynski2020}.
	& \MI
	\Tstrut \\ \hline

	\textbf{Invariance} 
	Let $\varphi : [0,1] \to [0,1]$ an increasing bijection.
	$$ I(x,y) = \varphi^{-1}(I(\varphi(x),\varphi(y)), \quad x,y \in [0,1].$$& \cite{Drewniak2006}
	& \IFI
	\Tstrut \\ \hline

	\textbf{} 
        $$I(x,y) \cdot I(y,z) = I(x,z)$$
        for all $x,y,z \in [0,1]$ such that $x>y>z$.
        & 
        This property is used in the characterization of $T$-power based implications \cite{Massanet2019B}.
	& 
	\Tstrut \\ \hline

	\textbf{Crispness} 
        $$I(x,y) \in \{0,1\}, \quad x,y \in [0,1]$$
        & 
        This property was imposed to study fuzzy implication functions which have a crisp domain \cite{Pinheiro2018}.
	& \C
	\Tstrut \\ \hline

	\textbf{Contrapositive symmetry with respect to a fuzzy negation $N$} 
        $$I(x,y) = I(N(y),N(x)), \quad x,y \in [0,1]$$
        & 
        Generalization of the classical tautology
        $$ p \to q \equiv \neg q \to \neg p$$
        \cite{Fodor1995}
	& \CPN
	\Tstrut \\ \hline

	\textbf{Law of left contraposition with respect to a fuzzy negation $N$} 
        $$I(N(x),y) = I(N(y),x), \quad x,y \in [0,1]$$
        & 
        
	& \LCPN
	\Tstrut \\ \hline

	\textbf{Law of right contraposition with respect to a fuzzy negation $N$} 
        $$I(x,N(y)) = I(y,N(x)), \quad x,y \in [0,1]$$
        & 
        
	& \RCPN
	\Tstrut \\ \hline

	\textbf{Law of importation with respect to a t-norm $T$} 
        $$I(T(x,y),z) = I(x,I(y,z)),\quad x,y,z \in [0,1]$$
        &  Generalization of the classical tautology
        $$(x \wedge y) \to z \equiv (x \to y) \to z$$
        It is used in the modification of the compositional rule of inference (CRI) called the hierarchical CRI which is more computational efficient \cite{Jayaram2008}.
	& \LI
	\Tstrut \\ \hline

	\textbf{Weak law of importation with respect to a function $F$}
 
 Let $F : [0,1]^2 \to [0,1]$ be a conjunctive, commutative and non-decreasing function.
        $$I(F(x,y),z) = I(x,I(y,z)),\quad x,y,z \in [0,1]$$
        &  To study the law of importation in a more general manner \cite{Massanet2011B}.
	& \WLI
	\Tstrut \\ \hline

	\textbf{Generalized law of importation}
 
        Let $C: [0,1]^2 \to [0,1]$ be a fuzzy conjunction, $I,J$ fuzzy implications and $\alpha \in (0,1)$
        $$I(C(x,\alpha),y) = I(x,J(\alpha,y)),\quad x,y,z \in [0,1]$$
        & To generalize the law of importation \cite{Baczynski2020}.
	& \GLI
	\Tstrut \\ \hline

	\textbf{Generalized cross-law of importation}
 
        Let $C: [0,1]^2 \to [0,1]$ be a fuzzy conjunction, $I,J$ fuzzy implications and $\alpha \in (0,1)$
        $$I(C(x,\alpha),y) = J(x,I(\alpha,y)),\quad x,y,z \in [0,1]$$
        & To generalize the law of importation \cite{Baczynski2020}.
	& \CLI
	\Tstrut \\ \hline

	\textbf{$T$-conditionality with respect to a t-norm $T$} 
        $$T(x,I(x,y)) 
        \leq y, \quad x,y \in [0,1]$$
        & 
         It is the generalization of the modus ponens
                 $$
        \displaystyle {\frac {P\to Q,P}{\therefore Q}}
        $$
        to fuzzy logic.
	& \TC
	\Tstrut \\ \hline

	\textbf{$O$-conditionality with respect to an overlap function $O$} 
        $$O(x,I(x,y)) 
        \leq y, \quad x,y \in [0,1]$$
        & 
         To generalize the $T$-conditionality using an overlap function \cite{Dimuro2019B}.
	& \OC
	\Tstrut \\ \hline

	\textbf{$U$-conditionality with respect to a uninorm $U$} 
        $$U(x,I(x,y)) 
        \leq y, \quad x,y \in [0,1]$$
        & 
         To generalize the $T$-conditionality using a uninorm \cite{MMas2019}.
	& \UC
	\Tstrut \\ \hline

	\textbf{$(T,N)$-Modus tollens with respect to a t-norm $T$ and a fuzzy negation $N$} 
        $$T(N(y),I(x,y)) \leq N(x), \quad x,y \in [0,1]$$
        & 
        It is the generalization of the modus tollens
        $$
        \displaystyle {\frac {P\to Q,\neg Q}{\therefore \neg P}}
        $$
        to fuzzy logic \cite{Trillas2005}.
	& \TNMT
	\Tstrut \\ \hline

	\textbf{Generalized hypothetical syllogism with respect to a t-norm $T$} 

        $$I(x,y) = \sup_{z \in [0,1]} T(I(x,z),I(z,y)), \quad x,y \in [0,1]$$
        & 
        It is the generalization of the hypothetical syllogism
                         $$
        \displaystyle {\frac {P\to Q,Q \to R}{\therefore P \to R}}
        $$
        to fuzzy logic \cite{Vemuri2017}. It plays an important role in approximate reasoning.
	& \GHS
	\Tstrut \\ \hline

	\textbf{Residuation principle with respect to a t-norm $T$} 
        $$T(x,z) \leq z \Leftrightarrow I(x,z) \geq z, \quad x,y,z \in [0,1]$$
        & 
        
	& \RP
	\Tstrut \\ \hline

	\textbf{$T$-transitivity with respect to a t-norm} 
        $$T(I(x,y),I(y,z)) \leq I(x,z), \quad x,y,z \in [0,1]$$
        & 
        \cite{Massanet2019B}.
	& \TT
	\Tstrut \\ \hline

	\textbf{Distributivity 1 with respect to a t-norm $T$ and a t-conorm $S$} 
        $$I(T(x,y),z) = S(I(x,z),I(y,z))$$ 
        $x,y,z \in [0,1]$
        & 
        Generalization of the classical tautology
        $$ (p \wedge q) \to r \equiv (p \to r) \vee (q \to r)$$
        \cite{Trillas2002}. The distribuitivity property is useful to avoid the combinatorial rule explosion in an inference mechanism \cite{Combs1998,Mendel1999}.
	& \DTS
	\Tstrut \\ \hline

	\textbf{Distributivity 2 with respect to a t-conorm $S$ and a t-norm $T$} 
        $$I(S(x,y),z) = T(I(x,z),I(y,z))$$ 
        $x,y,z \in [0,1]$
        & 
        Generalization of the classical tautology
        $$ (p \vee q) \to r \equiv (p \to r) \wedge (q \to r)$$
        \cite{Jayaram2004}. The distribuitivity property is useful to avoid the combinatorial rule explosion in an inference mechanism \cite{Combs1998,Mendel1999}.
	& \DST
	\Tstrut \\ \hline

	\textbf{Distributivity 3 with respect to two t-norms $T_1$ and $T_2$} 
        $$I(x,T_1(y,z)) = T_2(I(x,y),I(x,z))$$ 
        $x,y,z \in [0,1]$
        & 
        Generalization of the classical tautology
        $$ p \to (q \wedge r) \equiv (p\to q) \wedge (p \to r)$$
        \cite{Jayaram2004}. The distribuitivity property is useful to avoid the combinatorial rule explosion in an inference mechanism \cite{Combs1998,Mendel1999}.
	& \DTT
	\Tstrut \\ \hline

	\textbf{Distributivity 4 with respect to two t-conorms $S1$ and $S_2$} 
        $$I(x,S_1(y,z)) = S_2(I(x,y),I(x,z))$$ 
        $x,y,z \in [0,1]$
        & 
        Generalization of the classical tautology
        $$ p \to (q \vee r) \equiv (p\to q) \vee (p \to r)$$
        \cite{Jayaram2004}. The distribuitivity property is useful to avoid the combinatorial rule explosion in an inference mechanism \cite{Combs1998,Mendel1999}.
	& \DSS
	\Tstrut \\ \hline

	\textbf{Invariance with respect to the powers of a continuous t-norm $T$} 

        Let $r>0$
        $$I(x,y) = I\left(\xt{x}{r},\xt{y}{r}\right)$$ 
        $x,y \in (0,1)$ such that $\xt{x}{r},\xt{y}{r} \not = 0$.
        & 
        To impose that a fuzzy implication function which is used to model fuzzy conditionals should remain invariant when the same fuzzy hedges are used in antecedent and consequent (assuming that fuzzy hedges are modeled in terms of the powers of a continuous t-norm) \cite{Massanet2017}.
	& \PIT
	\Tstrut \\ \hline

	\textbf{Inverse invariance with respect to the powers of a continuous t-norm $T$} 

        Let $r>0$
        $$I(x,y) = I\left(\xt{x}{r},y_T^{\left( \frac{1}{r}\right)}\right)$$ 
        $x,y \in (0,1)$ such that $\xt{x}{r},y_T^{\left( \frac{1}{r}\right)} \not = 0$.
        & 
        It follows a similar reasoning than the $T$-power invariance but the consequent is modified using the quantifier inverse to the one used to modify the antecedent \cite{Baczynski2018}.
	& \PIIT
	\Tstrut \\ \hline

	\textbf{} 
        Let $N$ be a fuzzy negation
        $$I(x,N(x)) = N(x), \quad x \in [0,1].$$ 
        
        & 
        This property is valuable in fuzzy indices \cite{Bustince2003}.
	& 
	\Tstrut \\ \hline
\caption{List of additional properties of fuzzy implication functions alongside some comments about their underlying motivation and references.}\label{table:additional_properties}	
\end{longtable}

\section{Families of fuzzy implication functions}\label{sec:families}

In this section we focus on the current state of the art regarding the research on classes of fuzzy implication functions. This research line is motivated by the fact that, depending on the context and the proper rule and its behavior, various fuzzy implication functions with different properties can be adequate \cite{Trillas2008}. The most well-known families of fuzzy implication functions are the six ones collected in the surveys \cite{Mas2007,Baczynski2008B,Baczynski2015}: $(S,N)$-implications \cite{Trillas1985}, $R$-implications \cite{Trillas1985}, $QL$-implications \cite{Mas2006}, $D$-implications \cite{Mas2006}, and Yager's $f$ and $g$-implications \cite{Yager2004}. However, many other classes of fuzzy implication functions have been defined in recent years. According to the strategy used in the definition of a certain family, we can distinguish between four classes of fuzzy implication functions:

\begin{description}
	\item[S1.] \textbf{Classes generated from other fuzzy operators such as aggregation functions, fuzzy negations, etc.:} This strategy is based on the idea of combining adequately other fuzzy operators to obtain binary functions satisfying the axioms of the definition of a fuzzy implication function. Some of the most well-known classes such as $(S,N)$, $R$, $QL$, and $D$-implications belong to this strategy since they are generated by a t-conorm and a fuzzy negation; a t-norm; or a t-norm, a t-conorm and a fuzzy negation, respectively. More recently, other families like power-based implications \cite{Massanet2017}, Sheffer Stroke implications \cite{Baczynski2022B}, probabilistic and $S$-probabilistic implications \cite{Grzegorzewski2011}, or $(T,N)$-implications \cite{Bedregal2007} have been introduced also using this strategy.
	\item[S2.] \textbf{Classes generated from unary functions:} This strategy is based on the use of univalued functions (not necessarily fuzzy negations), often additive or multiplicative generators of other fuzzy logic connectives, to construct novel classes. These functions are usually called generators of the fuzzy implication function. This strategy experienced an important boost after Yager's $f$ and $g$-generated implications were introduced in \cite{Yager2004}.
	\item[S3.] \textbf{Classes generated from other fuzzy implication functions:} Adequately modifying the expression of already given fuzzy implication functions is another popular strategy to generate novel classes of these operators. This strategy has had an important revival lately and from the classical methods of the convex linear combination, the conjugation or the max/min construction (see \cite{Baczynski2008} for further details), more complex methods and especially, ordinal sums have recently appeared.
	\item[S4.] \textbf{Classes generated according to their final expression:} This strategy is based on fixing the desired final expression of these operators, and then studying when the corresponding functions fulfill the conditions in the definition of a fuzzy implication function. As compared with the other strategies, this one is quite new and it started in 2014, when polynomial implications were presented in \cite{Massanet2014} (see \cite{Massanet2022} for a deeper study on the polynomial implications).  
\end{description}
Apart from these four strategies, the proposal of generalizations of a certain class is quite popular, that is, to define a wider family which includes the original one. For instance, in \textbf{S1} the generalizations are usually based on considering a generalization of the fuzzy operators involved; or in \textbf{S2} they are based on weakening the conditions of the unary functions used or on generalizing the operator's expression. To express the relationship between a certain family and its generalizations, we will say that the generalizations are of the same ``type''. For example, we classify the generalizations of the $(S,N)$-implications as $(S,N)$ type implications. Having said this, intending to quantify the number of families introduced in the literature so far, in the subsequent sections we include a table with the name, expression, motivation and further comments of all the families gathered, separated according to classes \textbf{S1}-\textbf{S4} and several subclasses for the type of fuzzy operator used in the case of  \textbf{S1}.

\subsection{Basic fuzzy implication functions}

In Table \ref{table:basic_implications} examples of fuzzy implication functions can be found.

\begin{table}[h]
	\centering
	\begin{tabular}{|C{3cm}|l|} \hline
		\bf Name & \multicolumn{1}{|c|}{\bf Formula} \\   
		\hline \bf {\L}ukasiewicz & $\ILK(x,y)=\min\{1,1-x+y\}$ \\
		\hline \bf Gödel & $\IGD(x,y)=\left\{\begin{array}{ll}1&\hbox{if } x\leq y\\y&\hbox{if } x>y\end{array}\right.$ \\
		\hline \bf Reichenbach & $\IRC(x,y)=1-x+xy$  \\
		\hline \bf Kleene-Dienes & $\IKD(x,y)=\max\{1-x,y\}$ \\
		\hline \bf Goguen & $\IGG(x,y)=\left \{\begin{array}{ll} 1& \hbox{if } x\leq y\\ \frac{y}{x}&\hbox{if } x>y\end{array}\right.$ \\
		\hline \bf Rescher & $\IRS(x,y)=\left \{\begin{array}{ll} 1& \hbox{if } x\leq y\\ 0&\hbox{if } x>y\end{array}\right.$ \\
		\hline \bf Yager & $\IYG(x,y)=\left \{\begin{array}{ll} 1& \hbox{if } x=0 \hbox{ and } y=0\\ y^x&\hbox{if } x>0 \hbox{ or } y>0\end{array}\right.$ \\
		\hline \bf Weber & $\IWB(x,y)=\left \{\begin{array}{ll} 1& \hbox{if } x<1\\ y&\hbox{if } x=1\end{array}\right.$  \\
		\hline \bf Fodor & $\IFD(x,y)=\left \{\begin{array}{ll} 1& \hbox{if } x\leq y\\ \max\{1-x,y\}&\hbox{if } x>y\end{array}\right.$  \\
		\hline \bf Least & $\ILT(x,y)=\left\{\begin{array}{ll} 1&\hbox{if } x=0 \hbox{ or } y=1\\ 0&\hbox{if } x>0 \hbox{ and } y<1 \end{array}\right.$ \\
		\hline \bf Greatest & $\IGT(x,y)=\left\{\begin{array}{ll} 1&\hbox{if } x<1 \hbox{ or } y>0\\ 0&\hbox{if } x=1 \hbox{ and } y=0 \end{array}\right.$ \\
		\hline
	\end{tabular}
	\caption{Basic Fuzzy Implication Functions.}\label{table:basic_implications}
\end{table}

\newpage

\subsection{Families generated from other fuzzy operators}

\subsubsection{Families generated from fuzzy negations}
In Table \ref{table:families_negations} the reader can find the families of fuzzy implication functions generated from fuzzy negations.

\begin{longtable}{|p{0.45\textwidth}|p{0.45\textwidth}|}
	\hline
	\textbf{Name and Expression} & \textbf{Motivation, comments and references}   \Tstrut \\ \hline
    \textbf{$\mathbb{I}^{\mathbb{N}_S}$}
    $$
    I(x,y) = \left\{
        \begin{array}{ll}
            1 &  x \leq y, \\
            N(x) &  x>y,
        \end{array}
    \right.
    $$
    where $N$ is a fuzzy negation.
    &  Then main objective is to study fuzzy implication functions that satisfy \GHS \cite{Vemuri2017}. In this paper, the authors restricts his study to some of the well-known families of fuzzy implication functions, but since there do not exist many of them satisfying \GHS with respect to the minimum t-norm from these families, the authors proposes two new classes of fuzzy implication functions that do satisfy it. \Tstrut \\ \hline

        \textbf{$\mathbb{I}_{\mathbb{N}_S}$}
    $$
    I(x,y) = \left\{
        \begin{array}{ll}
            1 &  y=1, \\
            N(x) &  y<1,
        \end{array}
    \right.
    $$
    where $N$ is a fuzzy negation.
    & Then main objective is to study fuzzy implication functions that satisfy \GHS \cite{Vemuri2017}. In this paper, the authors restricts his study to some of the well-known families of fuzzy implication functions, but since there do not exist many of them satisfying \GHS with respect to the minimum t-norm from these families, the authors proposes two new classes of fuzzy implication functions that do satisfy it. \Tstrut \\ \hline


            \textbf{$f$-generated implications}
    $$
    I_f(x,y) = f^{-1}(f(N(x))f(y))
    $$
    where $N$ is a fuzzy negation and $f:[0,1] \to [0,1]$ is a continuous, strictly decreasing function with $f(1)=0$ and $f(0)=1$.
    &  The aim is to find a new method of generating fuzzy implication functions \cite{Souliotis2018}. 
    
    \textit{Remark:} Since $f$ is a fuzzy negation, if we consider $S$ the $f$-dual of the product t-norm we find out that the $f$-generated implications are a sub-family of the $(S,N)$-implications.\Tstrut \\ \hline


    \textbf{Neutral special implications}
    $$
    I(x,y) = \left\{
        \begin{array}{ll}
            1 & x \leq y, \\
            y + \frac{N(x-y)(1-x)}{1-x+y} & x>y,
        \end{array}
    \right.
    $$
    where $N$ is a fuzzy negation.
    & To introduce a family of fuzzy implication functions such that they satisfy \SPP \cite{Jayaram2009}. \Tstrut \\ \hline

    \textbf{$I^N$-implications}
    $$
    I(x,y) = \left\{
        \begin{array}{ll}
            1 & x \leq y, \\
            \frac{(1-N(x))y}{x} & x>y,
        \end{array}
    \right.
    $$
    where $N$ is a fuzzy negation.
    & This family appeared while the authors were studying the dependencies and independencies of several fuzzy implication properties \cite{Shi2010b}. \Tstrut \\ \hline

\caption{Families of fuzzy implication functions generated from fuzzy negations.}\label{table:families_negations}
\end{longtable}

\subsubsection{$(S,N)$-implications and generalizations}

In Table \ref{table:(s,n)-implications} the reader can find the families of fuzzy implication functions which are $(S,N)$-implications and generalizations.

\begin{longtable}{|p{0.55\textwidth}|p{0.45\textwidth}|}
	\hline
	\textbf{Name and Expression} & \textbf{Motivation, comments and references}   \Tstrut \\ \hline
    \textbf{$(S,N)$-implications} 
    $$
    I(x,y) = S(N(x),y),
    $$
    where $N$ is a fuzzy negation and $S$ is a t-conorm.
    &  \cite{Baczynski2008} To generalize the material implication of classical logic
    $$ p \to q \equiv \neg p \vee q.$$
    \Tstrut \\ \hline

    \textbf{$(U,N)$-implications} 
    $$
    I(x,y) = U(N(x),y),
    $$
    where $N$ is a fuzzy negation and $U$ is a disjunctive uninorm.
    &  To generalize $(S,N)$-implications using disjunctive uninorms \cite{DeBaets1999}.

    This family is suitable to define a fuzzy morphology based on uninorms.
    \Tstrut \\ \hline

        \textbf{$(Q,N)$-implications} 
    $$
    I(x,y) = Q(N(x),y),
    $$
    where $N$ is a fuzzy negation and $Q$ is a disjunctor.
    &  To generalize $(S,N)$-implications using disjunctors \cite{Yager2006}.

    The authors particularly study the case when $Q$ is a co-copula.
    \Tstrut \\ \hline

        \textbf{$(G_{f,N},N)$-implications} 
    $$
    I(x,y) = G(N(x),y),
    $$
    where $N$ is a strong, fuzzy negation and $G = \langle f,N \rangle$ is a DRAF.
    &  To generalize $S$-implications using DRAFs (Dual Representable Aggregation Functions), which are non-associative generalizations of nilpotent t-conorms \cite{Aguilo2010}.
    \Tstrut \\ \hline

        \textbf{$(TS,N)$-implications} 
    $$
    I(x,y) = f^{-1}((1-\lambda)f(T(N(x),y))+\lambda f(S(N(x),y))),
    $$
    where $N$ is a fuzzy negation, $T$ is a t-norm $S$ is a t-conorm, $\lambda \in [0,1]$ and $f:[0,1] \to \mathbb{R}$ is a continuous and strictly monotone function.
    &  To generalize $(S,N)$-implications using $TS$-functions \cite{Pradera2011}.
    \Tstrut \\ \hline

        \textbf{$(A,N)$-implications} 
    $$
    I(x,y) = A(N(x),y),
    $$
    where $N$ is a fuzzy negation and $A$ is an aggregation function
    &  To study necessary and sufficient conditions that have to fulfill an aggregation function $A$ so that the corresponding $(A,N)$-operator satisfies a certain property \cite{Ouyang2012}.

    \textit{Remark:} $(A,N)$-implications, although they appear as the general form of the classic material implication in fuzzy logic, they are actually another representation of all fuzzy implication functions. That is to say, every fuzzy implication function can be rewritten as an $(A,N)$-implication for some disjunctive aggregation function $A$ and a strong negation $N$ \cite{Pradera2016}.
    \Tstrut \\ \hline

        \textbf{Generalized $(S,N)$-implications} 
    $$
    I(x,y) = S((N(x))^{[n]}_S,y),
    $$
    where $n \in \mathbb{N}$, $N$ is a fuzzy negation and $S$ is a t-conorm.
    &  In \cite{VemuriJayaram2012} the authors present different operations between fuzzy implication functions with a view to propose novel generation methods to obtain a new fuzzy implication function from given ones. For each operation, the algebraic structure imposed by them on the set of all fuzzy implication functions is studied.
    \Tstrut \\ \hline

        \textbf{$(G,N)$-implications} 
    $$
    I(x,y) = G(N(x),y),
    $$
    where $N$ is a fuzzy negation and $G$ is a grouping function.
    &  To generalize $(S,N)$-implications using an aggregation function which is not necessarily associative \cite{Dimuro2014}.
    \Tstrut \\ \hline

        \textbf{$(U_S,N)$-implications} 
    $$
    I(x,y) = U_S(N(x),y),
    $$
    where $N$ is a fuzzy negation and $U_S$ is a disjunctive semi-uninorm.
    &  To generalize $(U,N)$-implications using disjunctive semi-uninorms $U_S$ \cite{Li2015}.
    \Tstrut \\ \hline

        \textbf{$(U_{CS},N)$-implications} 
    $$
    I(x,y) = U_{CS}(N(x),y),
    $$
    where $N$ is a fuzzy negation and $U_{CS}$ is a commutative, disjunctive semi-uninorm.
    &  To generalize $(U,N)$-implications using commutative, disjuntive semi-uninorms \cite{Li2015B}.
    \Tstrut \\ \hline

        \textbf{$(U_{P},N)$-implications} 
    $$
    I(x,y) = U_{P}(N(x),y),
    $$
    where $N$ is a fuzzy negation and $U_{P}$ is an associative disjunctive semi-uninorm.
    & To generalize $(U,N)$-implications using associative, disjunctive semi-uninorms \cite{Li2015B}.
    \Tstrut \\ \hline

        \textbf{$(U^2,N)$-operations} 
    $$
    I(x,y) = U^2(N(x),y),
    $$
    where $N$ is a fuzzy negation and $U^2$ is a disjunctive 2-uninorm.
    & To generalize $(U,N)$-implications using disjunctive 2-uninorms, i.e., uninorms with two neutral elements \cite{Zhou2020}.
    \Tstrut \\ \hline


            \textbf{New $(A,N)$-operations} 
    $$
    I(x,y) = g^{-1}(\min \{g(1),g(N(x) \wedge y) - g(N(N(x) \vee y))+ g(1)),
    $$
    where $N$ is a fuzzy negation.
    & To construct a subfamily of $(S,N)$-implications \cite{Peng2020}.
    \Tstrut \\ \hline
\caption{Families of fuzzy implication functions which are $(S,N)$-implications and generalizations.}\label{table:(s,n)-implications}
\end{longtable}

\subsubsection{$(T,N)$-implications and generalizations}

In Table \ref{table:(t,n)-implications} the reader can find the families of fuzzy implication functions which are $(T,N)$-implications and generalizations.

\begin{longtable}{|p{0.45\textwidth}|p{0.45\textwidth}|}
	\hline
	\textbf{Name and Expression} & \textbf{Motivation, comments and references}   \Tstrut \\ \hline


        \textbf{$(T,N)$-implications} 
        $$ I(x,y) = N(T(x,N(y))),$$
        where $N$ is a fuzzy negation and $T$ is a t-norm.
        & To define a new fuzzy implication function using a fuzzy negation and a t-norm in order to define an implicative De Morgan system $\langle T, N, I \rangle$ \cite{Bedregal2007}. Although these systems are equivalent to the De Morgan systems $\langle T, S, N \rangle$ where the t-conorm $S$ is defined by $T$ and $N$, the implicative De Morgan system defined have a natural and simple characterization.
        
        Notice that these fuzzy implication functions are a generalization of the following tautology
        $$ p \to q \equiv \neg (p \wedge \neg q).$$
        \Tstrut \\ \hline


        \textbf{$I_A$} 
        $$ I(x,y) = N(A(x,N(y))),$$
        where $N$ is a fuzzy negation and $A$ is an aggregation function with $A(1,0)=A(0,1)=0$.
        & To define and analyze a new fuzzy implication function obtained from overlap functions \cite{Zapata2014}. Indeed, the authors particularly studied the fuzzy implication functions given by
        $$ I(x,y) = N_1(G_0(x,N_2(y))),$$
        where $N_1$ and $N_2$ are strong fuzzy negations and $G_0$ is a grouping function.
        \Tstrut \\ \hline

        \textbf{$(U,f,g)$-implications} 
        $$ I(x,y) = g(U(x,f(y))),$$
        where $f$, $g$ are fuzzy negations and $U$ is a conjunctive uninorm.
        & To do a generalization of $(T,N)$-implications by making use of conjunctive uninorms and two different fuzzy negations. The authors also remark that these fuzzy implication functions are a generalization of $f$-generated implications \cite{Hlinena2014}.
        \Tstrut \\ \hline
\caption{Families of fuzzy implication functions which are $(T,N)$-implications and generalizations.}\label{table:(t,n)-implications}
\end{longtable}

\subsubsection{$R$-implications and generalizations}

In Table \ref{table:r-implications} the reader can find the families of fuzzy implication functions which are $R$-implications and generalizations.

\begin{longtable}{|p{0.45\textwidth}|p{0.45\textwidth}|}
	\hline
	\textbf{Name and Expression} & \textbf{Motivation, comments and references}   \Tstrut \\ \hline
        \textbf{$R$-implications}

        $$ I(x,y)
        =
        \sup \{ z \in [0,1] \mid T(x,z) \leq y\},$$
        where $T$ is a t-norm. & To generalize boolean implications \cite{Baczynski2008}.\Tstrut \\ \hline


        \textbf{$RU$-implications}

        $$ I(x,y)
        =
        \sup \{ z \in [0,1] \mid U(x,z) \leq y\},$$
        where $U$ is a uninorm. & To generalize $R$-implications using uninorms \cite{DeBaets1999}.
        
        Some further studies: idemptotent uninorms \cite{Ruiz2004}; uninorms continuous in $(0,1)^2$ \cite{Ruiz2009}; uninorms left-continuous, representable, continuous in $(0,1)^2$ or idempotent \cite{Aguilo2010}. \Tstrut \\ \hline

        \textbf{$G$-implications}
        
         $$I(x,y) = \sup \{ z \in [0,1] \mid G(x,z) \leq y\},$$
        where $G$ is a conjunctor. & To generalize $R$-implications using conjunctors \cite{Yager2006}.
        
        Some further studies: idemptotent uninorms \cite{Ruiz2004}; uninorms continuous in $(0,1)^2$ \cite{Ruiz2009}; uninorms left-continuous, representable, continuous in $(0,1)^2$ or idempotent \cite{Aguilo2010}. \Tstrut \\ \hline

        \textbf{$R$-implications from copulas and quasi-copulas}
        
         $$I(x,y) = \sup \{ z \in [0,1] \mid C(x,z) \leq y\},$$
        where $C$ is a certain conjunctor. & To generalize $R$-implications using different types of conjunctors: left-continuous semi-copulas, left-continuous pseudo t-norms, quasi-copulas and associative copulas \cite{Durante2007}. \Tstrut \\ \hline

        \textbf{Generalized $R$-implications}
        
         $$I(x,y) = \sup \{ z \in [0,1] \mid T(x_T^{[n]},z) \leq y\},$$
        where $n \in \mathbb{N}$ and $T$ is a t-norm. & In \cite{VemuriJayaram2012} the authors present different operations between fuzzy implication functions with a view to propose novel generation methods to obtain a new fuzzy implication function from given ones. For each operation, the algebraic structure imposed by them on the set of all fuzzy implication functions is studied. \Tstrut \\ \hline


        \textbf{$F_{g,N}$-implications}    
         $$I(x,y) = \sup \{ z \in [0,1] \mid F_{g,N}(x,z) \leq y\},$$
        where $F_{g,N}$ is a RAF. & To generalize $R$-implications using Representable Aggregation Functions (RAFs) \cite{Carbonell2010}. \Tstrut \\ \hline


        \textbf{$R$-implications from aggregation functions}    
         $$I(x,y) = \sup \{ z \in [0,1] \mid A(x,z) \leq y\},$$
        where $A$ is an aggregation function. & To study the necessary and sufficient conditions that has to fulfill an aggregation function $A$ so that the corresponding $R$-operator satisfies a certain property \cite{Ouyang2012}. \Tstrut \\ \hline


        \textbf{$R$-implications derived from semi-uninorms}    
         $$I(x,y) = \sup \{ z \in [0,1] \mid U(x,z) \leq y\},$$
         $$I(x,y) = \sup \{ z \in [0,1] \mid U(z,x) \leq y\},$$
        where $U$ is a semi-uninorm. & To generalize $R$-implications using semi-uninorms \cite{Liu2012}. \Tstrut \\ \hline


        \textbf{$R$-implications derived from fuzzy negations}    
         $$I(x,y) = \sup \{ z \in [0,1] \mid F_N(x,z) \leq y\},$$
        where $F_N$ is a semicopula defined as
        $$F_N(x,y) = \max \{0,x\wedge y - N(x \vee y)\},$$
        and $N$ is a fuzzy negation.
        & To generalize $R$-implications using semi-uninorms \cite{Liu2012}. \Tstrut \\ \hline


        \textbf{$R$-implications derived from overlap functions}    
         $$I(x,y) = \sup \{ z \in [0,1] \mid O(x,z) \leq y\},$$
        where $O$ is an overlap function.
        & To generalize $R$-implications using overlap functions \cite{Aguilo2013}. \Tstrut \\ \hline

\caption{Families of fuzzy implication functions which are $R$-implications and generalizations.}\label{table:r-implications}
\end{longtable}

\subsubsection{$QL$-operators and generalizations}

In Table \ref{table:ql-implications} the reader can find the families of fuzzy implication functions which are $QL$-implications and generalizations.

\begin{longtable}{|p{0.45\textwidth}|p{0.45\textwidth}|}
	\hline
	\textbf{Name and Expression} & \textbf{Motivation, comments and references}   \Tstrut \\ \hline
                \textbf{$QL$-operators}    
         $$I(x,y) = S(N(x),T(x,y)),$$
        where $N$ is a fuzzy negation, $S$ is a t-conorm and $T$ is a t-norm.
        &  To generalize the implication
        $$p \to q \equiv \neg p \vee (p \wedge q),$$
        defined in quantum logic \cite{Baczynski2008}.
        \Tstrut \\ \hline

        \textbf{$QLU$-operators}    
         $$I(x,y) = U_1(N(x),U_2(x,y)),$$
        where $N$ is a fuzzy negation and $U_1,U_2$ are uninorms. &
        To generalize $QL$-operators using uninorms \cite{Aguilo2013}.
        \Tstrut \\ \hline

        \textbf{$(O,G,N)$-operators}    
         $$I(x,y) = G(N(x),O(x,y)),$$
        where $N$ is a fuzzy negation, $G$ is a grouping function and $O$ is an overlap function. &
        To generalize $QL$-operators using overlap and grouping functions \cite{Dimuro2017}.
        \Tstrut \\ \hline
\caption{Families of fuzzy implication functions which are $QL$-implications and generalizations.}\label{table:ql-implications}
\end{longtable}

\subsubsection{$D$-operators and generalizations}

In Table \ref{table:d-implications} the reader can find the families of fuzzy implication functions which are $D$-implications and generalizations.

\begin{longtable}{|p{0.45\textwidth}|p{0.45\textwidth}|}
	\hline
	\textbf{Name and Expression} & \textbf{Motivation, comments and references}   \Tstrut \\ \hline
                \textbf{$D$-operators}    
         $$I(x,y) = S(y,T(N(x),N(y))),$$
        where $N$ is a fuzzy negation, $S$ is a t-conorm and $T$ is a t-norm.
        &  To generalize the Dishkant arrow
        $$ p \to q \equiv q \vee (\neg p \wedge \neg q),$$
        \cite{Baczynski2008}.
        \Tstrut \\ \hline

                \textbf{$DU$-operators}    
         $$I(x,y) = U_1(y,U_2(N(x),N(y))),$$
        where $N$ is a fuzzy negation and $U_1,U_2$ are uninorms.
        &  To generalize $D$-operators using uninorms \cite{Mas2007}.
        \Tstrut \\ \hline

                \textbf{$DU$-operators obtained from $(O,G,N)$ tuples}    
         $$I(x,y) = G(O(N(x),N(y)),y),$$
        where $N$ is a fuzzy negation, $O$ is an overlap function and $G$ is a grouping function.
        &  To generalize $D$-operators using overlap and grouping functions \cite{Dimuro2019}.
        \Tstrut \\ \hline

\caption{Families of fuzzy implication functions which are $D$-implications and generalizations.}\label{table:d-implications}
\end{longtable}

\subsubsection{Implications derived from copulas}

In Table \ref{table:impl_copulas} the reader can find the families of fuzzy implication functions derived from copulas.

\begin{longtable}{|p{0.4\textwidth}|p{0.5\textwidth}|}
	\hline
	\textbf{Name and Expression} & \textbf{Motivation, comments and references}   \Tstrut \\ \hline

        \textbf{Probabilistic implication}
        
        $$I(x,y) = \left\{
        \begin{array}{ll}
            1 & x=0, \\
            \frac{C(x,y)}{x} & x>0,
        \end{array}
    \right.$$
    where $C$ is a copula with $C(u_1,v)u_2 \geq C(u_2,v)u_1$ for every $u_1,u_2 \in [0,1]$ such that $u_1 \leq u_2$.
    &  To define a fuzzy implication function based on copulas combining fuzzy concepts and probability theory \cite{Grzegorzewski2011}. This family of operators seems to be useful in situations when we have to cope with imperfect knowledge which abounds with both kinds of uncertainty: imprecision and randomness. The idea beyond the definition of these fuzzy implication functions is first to interpret the probability of an implication as the conditional probability
    $$P(B\mid A) = \frac{P(A \cap B)}{P(A)},$$
    and then use the Sklar theorem to transform the problem into the unit square.  \Tstrut \\ \hline

        \textbf{Probabilistic $S$-implications}
        $$I(x,y) = C(x,y) -x+1,$$
        where $C$ is a copula.
    &  In order to propose a different approach to interpret the classical implication in probability theory, the material implication is considered \cite{Grzegorzewski2013}. According to this framework, the material implication interprets the probability of an implication as the probability that either $B$ occurs or $A$ does not occurs, i.e., $P(\neg A \cup B) = P(\neg A) + P(A \cap B)$. Then, the expression of this family of fuzzy implication functions is obtained by considering a copula and the Sklar theorem.  \Tstrut \\ \hline

        \textbf{Survival implications}
                $$I(x,y) = \left\{
        \begin{array}{ll}
            1 & x=0, \\
            \frac{x+y-1+C(1-x,1-y)}{x} & x>0,
        \end{array}
    \right.$$
        where $C$ is a copula with $C(1-u_1,1-v)u_2 - C(1-u_2,1-v)u_1 \geq (1-v)(u_2-u_1)$ for all $u_1,u_2 \in [0,1]$ such that $u_1 \leq u_2$.
    &  The definition of this family follows a similar approach as probabilistic implications but, in this case, the conditional probability considered is
    $$
    P(Y >y \mid X >x) = \frac{P(X>x,Y.Y)}{p(x>X)},
    $$
    where $X$ and $Y$ are random variables. This conditional probability depends on survival functions and according to Sklar's theorem, it can be considered in the unit square using a survival copula.

    \textit{Remark:} In \cite{Massanet2019D} it is proved that the families of probabilistic and survival implications are equivalent.
    \Tstrut \\ \hline

        \textbf{Survival $S$-implications}
                $$I(x,y) = y+C(1-x,1-y),$$
        where $C$ is a copula.
    &  Similarly to probabilistic $S$-implications, the perspective of the material implication and the probability
    $$
    P(X \leq Y \text{ or } Y>y) = P(X>x,Y>y)-P(X>x)+1,
    $$
    are considered.

    \textit{Remark:} In \cite{Massanet2019D} it is proved that the families of probabilistic and survival $S$-implications are equivalent.
    \Tstrut \\ \hline

        \textbf{Conditional implication}
                        $$I(x,y) = \left\{
        \begin{array}{ll}
            1 & x=0, \\
            \frac{\partial^*}{\partial x}C(x,y) & x>0,
        \end{array}
    \right.$$
        where $C$ is a copula and $\frac{\partial^*}{\partial x}C(x,y)$ is the extension of the partial derivative of a copula to every point in $[0,1]$ and $\frac{\partial^*}{\partial x}C(x,y)$ fulfills a certain condition.
    &  To propose a new family of fuzzy implication functions based on the conditional version of a copula. Specifically, the authors consider the following equality
    $$P(Y \leq y \mid X=x) = \frac{\partial}{\partial u} C(u,v),$$
    where the function $\frac{\partial}{\partial u} C(u,v)$ is defined almost everywhere. By defining an extension of the partial derivative of a copula to $[0,1]$ they define a new family of fuzzy implication functions.
    \Tstrut \\ \hline
\caption{Families of fuzzy implication functions generated from copulas.}\label{table:impl_copulas}
\end{longtable}

\subsubsection{Fuzzy implication functions related the power invariance property}

In Table \ref{table:power_invariant_impl} the reader can find families of fuzzy implication functions related to \PIT. 

\begin{longtable}{|p{0.45\textwidth}|p{0.45\textwidth}|}
	\hline
	\textbf{Name and Expression} & \textbf{Motivation, comments and references}   \Tstrut \\ \hline

        \textbf{Power-based implications}
        $$ I(x,y)
        = \sup \{ r \in [0,1] \mid y_T^{(r)} \geq x \},$$
        where $T$ is a continuous t-norm. & To define a family that satisfies \PIT with respect to a continuous t-norm.\cite{Massanet2017}.   \Tstrut \\ \hline

        \textbf{Strict $T$-power invariant implications}
        
        Let $T$ be a strict t-norm and $t$ an additive generator of $T$. Let $f:(0,1) \to [0,1]$ be a decreasing function and $\varphi : [0,+\infty] \to [0,1]$, $g:(0,1) \to [0,1]$  increasing functions such that $ \varphi(0)=0$,  $\varphi(+\infty)=1$ and 
	\begin{equation}
		\displaystyle\inf_{w \in (0,+\infty)} \varphi (w) \geq \max \left\lbrace \sup_{y \in (0,1)} g(y), \sup_{x \in (0,1)} f(x) \right\rbrace.
		\label{eq:strict:TPowerInv:MonotonicityCond}
	\end{equation}
	The function $I:[0,1]^2 \to [0,1]$ defined by
	\begin{equation}
		I(x,y) =\left\{ \begin{array}{ll}
			f(x) &   \text{if }   x \in (0,1) \text{ and } y=0, \\
			g(y) &  \text{if }  x = 1 \text{ and } y\in (0,1), \\
			\varphi \left(\frac{t(x)}{t(y)}\right) &  \text{otherwise},
		\end{array}
		\right.
		\label{eq:strict:TPowerInv:Expression}
	\end{equation}
	with the understanding $\frac{0}{0}=\frac{+\infty}{+\infty}=+\infty$, is called a strict $T$-power invariant implication.
         & This family is the characterization of all fuzzy implication functions that satisfy \PIT with respect to a strict t-norm $T$ \cite{Fernandez-Peralta2021}.   \Tstrut \\ \hline

        \textbf{Nilpotent $T$-power invariant implications}
        
        	Let $T$ be a nilpotent t-norm and $t$ an additive generator of $T$. Let $f:(0,1) \to [0,1]$ be a decreasing function and $\varphi:[0,+\infty] \to [0,1]$, $g: (0,1) \to [0,1]$ increasing functions such that $\varphi(0)=0$, $\varphi(+\infty)=1$ and
	$$
		f(x) \leq \inf_{y \in (0,1)} \varphi \left(\frac{t(x)}{t(y)}\right), \quad \text{for all } x \in (0,1),$$
		$$g(y) \leq \inf_{x \in (0,1)} \varphi \left(\frac{t(x)}{t(y)}\right), \quad \text{for all } y \in (0,1).$$
	The function $I:[0,1]^2 \to [0,1]$ defined  by
	\begin{equation}\label{eq:nilpot:TPowerInv:Expression}
		I =\left\{ \begin{array}{ll}
			1 & \text{if } x=0 \text{ and } y \in [0,1),\\
			f(x) &   \text{if }   x \in (0,1) \text{ and } y=0, \\
			g(y) &  \text{if }  x = 1 \text{ and } y\in (0,1), \\
			\varphi \left(\frac{t(x)}{t(y)}\right) &  \text{otherwise},
		\end{array}
		\right.
	\end{equation}
	with the understanding $\frac{0}{0} = + \infty$, is called a \emph{nilpotent $T$-power invariant implication}.
         & This family is the characterization of all fuzzy implication functions that satisfy \PIT with respect to a nilpotent t-norm $T$ \cite{Fernandez-Peralta2022}.   \Tstrut \\ \hline
\caption{Families of fuzzy implication functions related to \PIT.}\label{table:power_invariant_impl}
\end{longtable}

\subsubsection{Sheffer Stroke implications}

In Table \ref{table:shefferstroke_impl} the reader can find the families of fuzzy implication functions related to Sheffer Stroke operators.

\begin{longtable}{|p{0.45\textwidth}|p{0.45\textwidth}|}
	\hline
	\textbf{Name and Expression} & \textbf{Motivation, comments and references}   \Tstrut \\ \hline
        
        \textbf{$SS_{pq}$-implications}
        $$
        I(x,y) = N(T(x,N(T(x,y)))),
        $$
        where $N$ is a fuzzy negation and $T$ is a t-norm.
        & To define a new family of fuzzy implication functions considering fuzzy Sheffer stroke operators and the two possible expressions of the classical implication in terms of the classical sheffer stroke operator \cite{Baczynski2022B}.   \Tstrut \\ \hline

        \textbf{$SS_{qq}$-implications}
        $$
        I(x,y) = N(T(x,N(T(y,y)))),
        $$
        where $N$ is a fuzzy negation and $T$ is a t-norm.
        & To define a new family of fuzzy implication functions considering fuzzy Sheffer stroke operators and the two possible expressions of the classical implication in terms of the classical sheffer stroke operator \cite{Baczynski2022B}.   \Tstrut \\ \hline
        
\caption{Families of fuzzy implication related to Sheffer Stroke operators.}\label{table:shefferstroke_impl}
\end{longtable}

\subsection{Families constructed according to their final expression}

In Table \ref{table:final_expression_impl} the reader can find the families of fuzzy implication functions that were constructed fixing their final expression.

\begin{longtable}{|p{0.45\textwidth}|p{0.45\textwidth}|}
	\hline
	\textbf{Name and Expression} & \textbf{Motivation, comments and references}   \Tstrut \\ \hline
        \textbf{Fuzzy polynomial implications} Let $n \in \mathbb{N}$
        $$I(x,y)
        =
        \sum_{\substack{0 \leq i,j \leq n \\ i+j \leq n}} a_{ij}x^iy^j,
        $$
        where $a_{ij} \in \mathbb{R}$ and there exist some $0 \leq i,j \leq n$ with $i+j=n$ such that $a_{ij} \not = 0$. & To introduce a family of fuzzy implication functions with a simple final expression from the computational point of view in order to be more robust to numerical computational errors \cite{Massanet2014}.
        \\ \hline
        \textbf{\OP-polynomial implications} Let $n \in \mathbb{N}$
        $$I(x,y)
        =
        \left\{
        \begin{array}{ll}
            1 & x \leq y, \\
            \displaystyle \sum_{\substack{0 \leq i,j \leq n \\ i+j \leq n}} a_{ij}x^iy^j, & x>y,
        \end{array}
        \right.
        $$
        where $a_{ij} \in \mathbb{R}$ and there exist some $0 \leq i,j \leq n$ with $i+j=n$ such that $a_{ij} \not = 0$. & Since fuzzy polynomial implications do not satisfy the ordering property \OP the authors propose a family of fuzzy implication functions satisfying this property and having in the rest of their domain a polynomial expression \cite{Massanet2015}.
        \\ \hline
        \textbf{Rational fuzzy implication functions} Let $n,m \in \mathbb{N}$
        $$I(x,y)
        = \frac{p(x,y)}{q(x,y)} = \frac{\displaystyle \sum_{\substack{0 \leq i,j \leq n \\ i+j \leq n}} a_{ij}x^iy^j}{\displaystyle \sum_{\substack{0 \leq s,t \leq n \\ s+t \leq n}} b_{st}x^sy^t},
        $$
        where 
        \begin{itemize}
        \item $a_{ij} \in \mathbb{R}$ and there exists some $0 \leq i,j \leq n$ with $i+j = n$ such that $a_{ij} \not = 0$.
        \item $b_{st} \in \mathbb{R}$ and there exist some $0 \leq s,t \leq m$ with $s+t=m$ such that $b_{st} \not = 0$.
        \item $p$ and $q$ have no factors in common.
        \item $q(x,y) \not = 0$ for all $x,y \in [0,1]$.
        \end{itemize}
        
        & Following the same reasoning than for fuzzy polynomial implications, the authors define rational fuzzy implication functions as those fuzzy implications whose expression is given by the quotient of two polynomials of two variables \cite{Massanet2016}.
        \\ \hline
\caption{Families of fuzzy implication functions constructed according to their final expression.}\label{table:final_expression_impl}
\end{longtable}

\begin{remark} A similar approach to the introduction of fuzzy polynomial implications is the consideration of parametrized fuzzy implications \cite{Whalen1996,Whalen2003}. In this approach, the author considers $S$ and $R$-implications generated from a parametrized family of t-conorms or t-norms, respectively.
\end{remark}

\subsection{Families generated from unary functions}
In Table \ref{table:unary_generated_impl} the reader can find the families of fuzzy implication functions generated from unary functions.

\begin{longtable}{|p{0.55\textwidth}|p{0.35\textwidth}|}
	\hline
	\textbf{Name and Expression} & \textbf{Motivation, comments and references}   \Tstrut \\ \hline
        \textbf{$I_f$ implications}
            $$
            I(x,y) = \left\{
        \begin{array}{ll}
            1 & x \leq y, \\
            f^{(-1)}(f(y^+)-f(x)) & x>y,
        \end{array}\right.$$ 
        
    where $f:[0,1] \to [0,+\infty)$ is a strictly decreasing, continuous function with $f(1^+)=f(1)=0$.& \cite{Biba2012}.\\  \hline

            \textbf{$I_N^g$ implications}
            $$I(x,y) = g^{(-1)}(g(N(x)) + g(y)),$$
    where $N$ is a fuzzy negation and $g:[0,1] \to [0,+\infty)$ is a strictly increasing, continuous function with $g(0)=0$.& \cite{Biba2012}.\\  \hline

        \textbf{$f$-implications}
    $$I(x,y) = f^{-1}(xf(y)),$$
    where $f:[0,1] \to [0,+\infty)$ is a strictly decreasing, continuous function with $f(1)=0$. In this case, $f$ is called an $f$-generator. & In \cite{Yager2004} the author proposes a novel method to define a family of fuzzy implication functions by making use of continuous additive generators of continuous Archimedean t-norms and t-conorms, respectively.
    
    In the same paper, the author gives an extensive analysis of the role of these new classes of implications in approximate reasoning. In particular, he introduces and studies some new interesting concepts like strictness of implications, sharpness of inference or the strictness index. According to him, these new fuzzy implication functions are interesting in approximate reasoning since they accomplish strictness of a fuzzy implication and sharpness of inference.
    \\  \hline

        \textbf{$g$-implications}
    $$I(x,y) = g^{(-1)}\left(\frac{1}{x}g(y)\right),$$
    where $g:[0,1] \to [0,+\infty)$ is a strictly increasing, continuous function with $g(0)=0$. In this case, $g$ is called an $g$-generator. & In \cite{Yager2004} the author proposes a novel method to define a family of fuzzy implication functions by making use of continuous additive generators of continuous Archimedean t-norms and t-conorms, respectively.
    
    In the same paper, the author gives an extensive analysis of the role of these new classes of implications in approximate reasoning. In particular, he introduces and studies some new interesting concepts like strictness of implications, sharpness of inference or the strictness index. According to him, these new fuzzy implication functions are interesting in approximate reasoning since they accomplish strictness of a fuzzy implication and sharpness of inference.
    \\  \hline

    \textbf{$h$-generated implications}
    $$I(x,y) = h^{(-1)}(xh(y)),$$
    where $h:[0,1] \to [0,1]$ is a strictly decreasing, continuous function with $h(0)=1$. & To follow a similar approach to the definition of Yager's implications but using multiplicative generators of continuous Archimedean t-conorms \cite{Jayaram2006}.

    \textit{Remark:} In \cite{Baczynski2007B} it is proved that they are a subfamily of $(S,N)$-implications.
    \\  \hline


    \textbf{Minimal special implications}
    $$ I(x,y) = f(x-y),$$
    where $f:[-1,1] \to [0,1]$ is a non-increasing function such that $f(y)=1$ if $y \leq 0$ and $f(1)=0$. & To define the minimal special implication with respect to a fuzzy negation \cite{Jayaram2009}. \\ \hline


        \textbf{$h$-implications}
                $$
            I(x,y) = \left\{
        \begin{array}{ll}
            1 & x=0, \\
            h^{(-1)}(xh(y)) & x>0 \text{ and } y \leq e, \\
            h^{-1}\left(\frac{1}{x}h(y)\right) & x>0 \text{ and } y > e,
        \end{array}\right.$$
    where $h:[0,1] \to (-\infty,+\infty)$ is a strictly increasing, continuous function with $h(e)=0$ and $h(1)=+\infty$. In this case, $h$ is called an $h$-generator. & To define a new family similar to Yager's implications but generated from additive generators of representable uninorms \cite{Baczynski2013,Massanet2011A}. \\ \hline

    \textbf{$(h,e)$-implications}
                $$
            I(x,y) = \left\{
        \begin{array}{ll}
            1 & x=0, \\
            h^{(-1)}\left(\frac{x}{e}h(y)\right) & x>0 \text{ and } y \leq e, \\
            h^{-1}\left(\frac{e}{x}h(y)\right) & x>0 \text{ and } y > e,
        \end{array}\right.$$
    where $h:[0,1] \to (-\infty,+\infty)$ is a strictly increasing, continuous function with $h(e)=0$ and $h(1)=+\infty$. In this case, $h$ is called an $h$-generator. & To modify the definition of $h$-implications to define a family that satisfies \NPe \cite{Massanet2011A}. \\ \hline


    \textbf{Generalized $f$-implications}
                $$
            I(x,y) = f^{-1}(h(x)f(y))$$
    where $f:[0,1] \to [0,+\infty)$ is a strictly decreasing, continuous function with $f(1)=0$ and $h:[0,1] \to [0,1]$ is a increasing bijection. & In \cite{VemuriJayaram2012} the authors present different operations between fuzzy implication functions with a view to propose novel generation methods to obtain a new fuzzy implication function from given ones. For each operation, the algebraic structure imposed by them on the set of all fuzzy implication functions is studied. \\ \hline

    
    \textbf{Generalized $g$-implications}
                $$
            I(x,y) = g^{(-1)}(k(x)g(y))$$
    where $g:[0,1] \to [0,+\infty)$ is a strictly increasing, continuous function with $g(0)=0$ and $k:[0,1] \to [1,+\infty)$ is a decreasing function with $k(1)=1$. & In \cite{VemuriJayaram2012} the authors present different operations between fuzzy implication functions with a view to propose novel generation methods to obtain a new fuzzy implication function from given ones. For each operation, the algebraic structure imposed by them on the set of all fuzzy implication functions is studied. \\ \hline


        \textbf{$f$-generated operations}
                $$
            I(x,y) = f^{-1}(F(x,f(y))),$$
    where $f:[0,1] \to [0,+\infty)$ is a strictly decreasing, continuous function with $f(1)=0$ and $F:[0,1]^2 \to [0,1]$ is a binary function. & To generalize Yager's implications by considering a more general internal function than the product into their expression \cite{Massanet2012C}.

    \textit{Remark:} In \cite{Hlinena2012} the authors introduce independently the $f$-generated operations (which they call $(T,f)$-implications) in the case when $F$ is a t-norm.
    
    \\ \hline


    \textbf{$g$-generated operations}
                    $$
            I(x,y) = g^{(-1)}\left(F\left(\frac{1}{x},g(y)\right)\right),$$
    where $g:[0,1] \to [0,+\infty)$ is a strictly increasing, continuous function with $g(0)=0$ and $F:[0,1]^2 \to [0,1]$ is a binary function. & To generalize Yager's implications by considering a more general internal function than the product into their expression \cite{Massanet2012C}.
    
    \\ \hline


    \textbf{$(*,F)$-implications}
                    $$
            I(x,y) = h^{(-1)}(g(y)-f(x)),$$
    where $f,g,h$ are increasing and non-constant functions from $[0,1]$ to $[-\infty,+\infty]$ such that
    \begin{itemize}
    \item $h$ is strictly increasing in a left-neighborhood of 1.
    \item $h(0^+) \geq g(0) -f(1)$.
    \item $h(1^-) \leq \min \{g(1)-f(1),g(0)-f(0)\}$.
    \end{itemize}
     & To generalize $I_f$ implications \cite{Hlinena2012}. \\ \hline


    \textbf{$(*,F,N)$-implications}
                    $$
            I(x,y) = h^{(-1)}(g(y)-f(N(x))),$$
    where $f,g,h$ are increasing and non-constant functions from $[0,1]$ to $[-\infty,+\infty]$ such that
    \begin{itemize}
    \item $h$ is strictly increasing in a left-neighborhood of 1.
    \item $h(0^+) \geq f(0)+g(0)$.
    \item $h(1^-) \leq \min \{f(1)+g(0),f(0)+g(1)\}$.
    \end{itemize}
     & To generalize $I_N^g$ implications \cite{Hlinena2012}. \\ \hline


    \textbf{$(g,\min)$-implications}
        $$
        I(x,y) =
            \left\{
        \begin{array}{ll}
            1 & x=0, \\
            y & g(1)>1,  1 \leq g(y) \leq g(x), \\
            g^{\ominus} \left( \min \{\frac{1}{x}g(y)\}\right) & \text{otherwise},
        \end{array}
    \right.$$
    where $g:[0,1] \to [0,+\infty)$ is a strictly increasing, continuous function with $g(0)=0$ and $g^{\ominus}$ its partial inverse. 
    
     & To introduce a new family of fuzzy implication functions following the same approach as Yager's implications by using additive generators of continuous, Archimedean t-conorms and partial inverses \cite{Liu2012}. \\ \hline


        \textbf{$(f,g)$-implications}
        $$
        I(x,y) = f^{(-1)}(g(x)f(y))$$
    where $f:[0,1] \to [0,+\infty)$ is a strictly decreasing, continuous function with $f(1)=0$ and $g:[0,1] \to [1,+\infty)$ is a strictly increasing, continuous function with $g(0)=0$.
    
     & To propose a generalization of Yager's $f$-implications by means of generalizing the internal factor $x$ to a more general unary function \cite{Massanet2013B}. However, the main motivation is that a certain subfamily of this one called $(f,e)$-implications is related to the characterization of $(h,e)$-implications \cite{Fernandez-Peralta2021B}. \\ \hline


        \textbf{$(g,f)$-implications}
        $$
        I(x,y) = g^{(-1)}(f(x)g(y))$$
    where $g:[0,1] \to [0,+\infty)$ is a strictly increasing, continuous function with $g(0)=0$ and $f:[0,1] \to [1,+\infty)$ is a strictly decreasing, continuous function with $f(0)=+\infty$.
    
     & To propose a generalization of Yager's $g$-implications by means of generalizing the internal factor $\frac{1}{x}$ to a more general unary function \cite{Massanet2013B}. However, the main motivation is that a certain subfamily of this one called $(g,e)$-implications is related to the characterization of $(h,e)$-implications \cite{Fernandez-Peralta2021B}. \\ \hline


        \textbf{$h^{-1}$-implications} There are two possible definitions:
        $$
        I(x,y) = \left\{
        \begin{array}{ll}
            1 & x=0, \\
            h^{-1}\left(\max\{-x,h(y)\}\right) & 0<x\leq y \leq e, \\
            y & \text{otherwise},
        \end{array}
    \right.$$
    where $e \in (0,1)$, $h:[0,1] \to (-\infty,+\infty)$ is a strictly increasing, continuous function with $h(e)=0$, $h(1) \leq 1$.  

            $$
        I(x,y) = \left\{
        \begin{array}{ll}
            1 & x=0 \text{ or } x \leq y, h(y) \geq 1, \\
            h^{-1}\left(\max\{-x,h(y)\}\right) & 0<x\leq y \leq e, \\
            y & \text{otherwise},
        \end{array}
    \right.$$
    where $e \in (0,1)$, $h:[0,1] \to (-\infty,+\infty)$ is a strictly increasing, continuous function with $h(e)=0$, $h(1) > 1$.
    
     & To follow a similar approach as the definition of Yager's implications and $h$-implications but using generalized additive generators of representable uninorms \cite{Liu2013}. \\ \hline


        \textbf{$(h,\min)$-implications}
        $$
        I(x,y) = \left\{
        \begin{array}{ll}
            1 & x=0, \\
            h^{\ominus}\left(\max\{-x,h(y)\}\right) & x>0, y \leq e, \\
            h^{\ominus}\left(\min\{\frac{1}{x},h(y)\}\right) & x>0, y >  e,
        \end{array}
    \right.$$
    where $h$ is an $h$-generator and $h^{\ominus}$ its partial inverse.
    
     & To introduce a new class of fuzzy implication functions following the same approach as $h$-implications by using generalized $h$-generators and partial inverses \cite{Liu2013B}. \\ \hline     


            \textbf{$(f,g,h)$-implications}
        $$
        I(x,y) = h(f(x)+g(y)),$$
        where $f,g,h$ are increasing, non-constant functions from $[0,1]$ to $[-\infty,+\infty]$ such that
        \begin{itemize}
        \item $f(0) > f(1)$.
        \item $g(0)<g(1)$.
        \item $h(f(1)+g(0))=0$.
        \item $h(\min\{f(0)+g(0),f(1)+g(1)\})=1$.
        \end{itemize}
    
     & To continue the study started in \cite{Hlinena2012} where the authors notice that in some cases fuzzy implication functions can be represented in the form $h(f(x)*g(y))$, where $*$ is one of the usual arithmetic operations and $f,g,h$ are monotone functions \cite{Hlinena2013B}. \\ \hline 


            \textbf{$(f,g)$-implications}
        $$
        I(x,y) = f^{(-1)}(g(x)f(y)),$$
        where $f:[0,1] \to [0,+\infty)$ is a strictly decreasing, continuous function with $f(1)=0$ and $g:[0,1] \to [0,1]$ is an increasing function satisfying $g(0)=0$ and $g(1)=1$.
    
     & To generalize Yager's $f$-implications \cite{XieLiu2013}. \\ \hline


            \textbf{$(g,u)$-implications}
        $$
        I(x,y) = g^{(-1)} \left(u \left(\frac{1}{x},g(y)\right)\right),$$
        where $g:[0,1] \to [0,+\infty)$ is a strictly increasing, continuous function with $g(0)=0$ and $u:[0,1] \times [0,g(1)] \to [0,+\infty)$ is a non-decreasing in each argument and satisfies $u(+\infty,0)=+\infty$ and $u(1,y)=y$ for all $y \in [0,g(1)]$.
    
     & To generalize Yager's $g$-implications \cite{ZhangLiu2013}. \\ \hline


                 \textbf{$(f,g,\wedge)$-implications}
        $$
        I(x,y)=I_0(x,y) \vee (f^{\shortdownarrow}(x) \wedge g^{\shortuparrow}(y) \wedge I_1(x,y)),$$
        where $f,g$ are to unary functions from $[0,1]$ to $[0,1]$,
        $$f^{\shortdownarrow}(x) = \sup\{f(u) \mid u \geq x\},$$
        $$g^{\shortuparrow}(x) = \sup\{g(u) \mid u \leq x\},$$
        $$
        I_0(x,y) = \left\{
        \begin{array}{ll}
            1 & x=0 \text{ or } y=1, \\
            x & \text{otherwise},
        \end{array}
    \right.
        $$
        $$
        I_1(x,y) = \left\{
        \begin{array}{ll}
            0 & x=1 \text{ and } y=0, \\
            1 & \text{otherwise}.
        \end{array}
    \right.
        $$
     & To introduce a new method of generating fuzzy implication functions from one-variable functions \cite{Su2015}. \\ \hline


                 \textbf{$(f,g,\vee)$-implications}
        $$
        I(x,y)=I_1(x,y) \vee (f^{\shortdownarrow}(x) \wedge g^{\shortuparrow}(y) \wedge I_0(x,y)),$$
        where $f,g$ are to unary functions from $[0,1]$ to $[0,1]$,
        $$f^{\shortdownarrow}(x) = \sup\{f(u) \mid u \geq x\},$$
        $$g^{\shortuparrow}(x) = \sup\{g(u) \mid u \leq x\},$$
        $$
        I_0(x,y) = \left\{
        \begin{array}{ll}
            1 & x=0 \text{ or } y=1, \\
            x & \text{otherwise},
        \end{array}
    \right.
        $$
        $$
        I_1(x,y) = \left\{
        \begin{array}{ll}
            0 & x=1 \text{ and } y=0, \\
            1 & \text{otherwise}.
        \end{array}
    \right.
        $$
     & To introduce a new method of generating fuzzy implication functions from one-variable functions \cite{Su2015}. \\ \hline


                      \textbf{Generalized $f$-implications}
        $$
        I(x,y)=f_2^{(-1)}(xf_1(y)),$$
        where $f_1, f_2$ are strictly decreasing, continuous functions from $[0,1]$ to $[0,+\infty)$ with $f_1(1)=f_2(1)=0$.
        
     & To generalize Yager's $f$-implications \cite{ZhangZhang2017}. \\ \hline


      \textbf{Generalized $g$-implications}
        $$
        I(x,y)=g_2^{(-1)}\left(\frac{1}{x}g_1(y)\right),$$
        where $g_1,g_2$ are strictly increasing, continuous functions from $[0,1]$ to $[0,+\infty)$ with $g_1(0)=g_2(0)=0$.
        
     & To generalize Yager's $g$-implications \cite{ZhangZhang2017}. \\ \hline

     \textbf{Generalized $g$-implications}
        $$
        I(x,y)=g^{(-1)}(f(x)g(y)),$$
        where $g:[0,1] \to [0,+\infty)$ is a strictly increasing, continuous function with $g(0)=0$ and $f:[0,1] \to [1,+\infty]$ is a decreasing, continuous function with $f(0)+\infty$, $f(1)=1$.
        
     & To generalize Yager's $g$-implications \cite{PeiZhu2017}. \\ \hline


     \textbf{$(h,f,g)$-implications}
        $$
        I(x,y)=\left\{
        \begin{array}{ll}
            1 & x=0, \\
            h^{(-1)}(f(x)h(y)) & x>0, y \leq e, \\
            h^{-1}(g(x)h(y)) & x>0, y > e,
        \end{array}
    \right.$$
    where $h:[0,1] \to (-\infty,+\infty)$ is a strictly increasing, continuous function with $h(e)=0$ and $h(1)=+\infty$, $f:[0,1] \to [0,1]$ is an increasing function with $f(0)=0$, $f(1)=1$ and $g:[0,1] \to [1,+\infty)$ is a decreasing function with $g(0)=+\infty$ and $g(1)=1$.
        
     & To generalize $h$-implications \cite{PeiZhu2017B}. \\ \hline


          \textbf{$\mathbb{S}$}
        $$
        I(x,y)=\left\{
        \begin{array}{ll}
            1 & y=1, \\
            \psi(x) & y <1,
        \end{array}
    \right.$$
    where $\psi:[0,1] \to [0,1]$ is an increasing function with $\psi(0)=0$ and $\psi(1)=1$.
     & In \cite{Vemuri2017} the main objective is to study fuzzy implication functions that satisfy \GHS, one of the main inference rules. In this article, the author restricts his study to some of the well-known families of fuzzy implication functions. Since there are not many fuzzy implication functions satisfying this property with respect to the minimum t-norm, the author proposes two new families of fuzzy implication functions satisfying \GHS with respect to the minimum t-norm.  \\ \hline


          \textbf{$\mathbb{I}_{\psi}$}
        $$
        I(x,y)=\left\{
        \begin{array}{ll}
            1 & x \leq y, \\
            \psi(x) & x>y,
        \end{array}
    \right.$$
    where $\psi:[0,1] \to [0,1]$ is an increasing function with $\psi(0)=0$ and $\psi(1)=1$.
     & In \cite{Vemuri2017} the main objective is to study fuzzy implication functions that satisfy \GHS, one of the main inference rules. In this article, the author restricts his study to some of the well-known families of fuzzy implication functions. Since there are not many fuzzy implication functions satisfying this property with respect to the minimum t-norm, the author proposes two new families of fuzzy implication functions satisfying \GHS with respect to the minimum t-norm.  \\ \hline


          \textbf{Preference implication}
        $$
        I(x,y)=\left\{
        \begin{array}{ll}
            1 & (x,y) \in \{(0,0),(1,1)\}, \\
            f^{-1}\left(f(\nu)\frac{f(y)}{f(x)}\right) & \text{otherwise},
        \end{array}
    \right.$$
    where $f$ is the generator of a strict t-norm and $\nu$ is the fixed point of the corresponding negation.
     & The only continuous fuzzy implication function which satisfies \EP, \OP, \NP is \ILK. The main motivation in \cite{Dombi2021B} is to solve the problem of finding non-trivial solutions to all possible distributivity equations. In this sense, the preference implication satisfies all four distributivity equations with respect to the operators of the pliant system. In addition, the preference implication is closely related to the preference relation used in multicriteria decision making.
     
     \textit{Remark}: \ILK satisfies the four distributivity equations only with the maximum and the minimum.
     \\ \hline


          \textbf{$(\theta,t)$-implications}
        $$
        I(x,y)= \theta^{(-1)}(\min\{t(x)+\theta(y),1\})
        $$
    
    where $\theta$ is a multiplicative generator of a t-norm, $t$ an additive generator of another (possibly the same) t-norm and $t(0) \geq \theta(1^-) - \theta(0)$.
     & To introduce a new family using two generator functions and the addition as the arithmetic operation (instead of the multiplication or division) \cite{Zhou2021}. Moreover, the author proposes a parallel hierarchical method base on \LI and \FOP (properties that are satisfied by the new family in some cases).\\ \hline

\caption{Families of fuzzy implication functions generated from unary functions.}\label{table:unary_generated_impl}
\end{longtable}

\subsection{Families generated from other fuzzy implication functions (also called construction methods)}\label{sec:construction_methods}

\begin{longtable}{|p{0.6\textwidth}|p{0.4\textwidth}|}
	\hline
	\textbf{Name and Expression} & \textbf{Motivation, comments and references}   \Tstrut \\ \hline

    \textbf{Upper contrapositivisation} 

    Let $N$ be a strong fuzzy negation,
    $$I_N^U(x,y) = \max\{I(x,y),I(N(y),N(x))\},$$
    for all $x,y \in [0,1]$.
    
    & The motivation is to modify a fuzzy implication function which may not satisfy \CPN in order to satisfy this property with respect to a strong fuzzy negation \cite{Bandler1980}. \Tstrut \\ \hline

    \textbf{Lower contrapositivisation} 

    Let $N$ be a strong fuzzy negation,
    $$I_N^L(x,y) = \min\{I(x,y),I(N(y),N(x))\},$$
    for all $x,y \in [0,1]$.
    
    & The motivation is to modify a fuzzy implication function which may not satisfy \CPN in order to satisfy this property with respect to a strong fuzzy negation \cite{Bandler1980}. \Tstrut \\ \hline

    \textbf{Medium contrapositivisation} 

    Let $N$ be a strict fuzzy negation,
    $$I_N^M(x,y) = \min\{I(x,y) \vee N(x),I(N(y),N(x))\vee y\},$$
    for all $x,y \in [0,1]$.
    
    & The motivation is to modify a fuzzy implication function which may not satisfy \CPN in order to satisfy this property, \NP and it is $N$-compatible independetly from the ordering of $N$ and $N_I$. \cite{Jayaram2006}. \Tstrut \\ \hline

    \textbf{$N$-lower contrapositivisation} 

    Let $N$ be a strong fuzzy negation,
    $$I_N^{LC}(x,y) = \left\{
        \begin{array}{ll}
            I(x,y) & y \geq N(x), \\
            I(N(y),N(x)) & y < N(x),
        \end{array}
    \right.$$
    for all $x,y \in [0,1]$.
    
    & To propose new types of contrapositivisation \cite{Aguilo2015}. \Tstrut \\ \hline

    \textbf{$N$-modified-lower contrapositivisation} 

    Let $N$ be a strong fuzzy negation,
    $$\tilde{I}_N^{LC}(x,y) = \left\{
        \begin{array}{ll}
            I(x,y) & \quad y > N(x), \\
            I(N(y),N(x)) & \quad y \leq N(x),
        \end{array}
    \right.$$
    for all $x,y \in [0,1]$.
    
    & To propose new types of contrapositivisation \cite{Aguilo2015}. \Tstrut \\ \hline


    \textbf{$N$-reciprocation} 

    Let $N$ be a fuzzy negation
    $$I_N(x,y) = I(N(y),N(x)),$$
    for all $x,y \in [0,1]$.
    
    & The motivation is to obtain a new fuzzy implication functions when the original one does to satisfy \CPN \cite[Definition 1.6.1]{Baczynski2008}. \Tstrut \\ \hline


    \textbf{$\varphi$-reciprocation} 

    Let $\varphi : [0,1] \to [0,1]$ be an increasing bijection
    $$I_\varphi(x,y) = \varphi^{-1}(I(\varphi(y),\varphi(x))),$$
    for all $x,y \in [0,1]$.
    
    & This methods preserves most of the additional properties of fuzzy implication functions and let us define conjugacy classes \cite{Baczynski2008}. \Tstrut \\ \hline


    \textbf{\IP-modification} 
    $$I^{(1)}(x,y) = \left\{
        \begin{array}{ll}
            1 & \quad x \leq y, \\
            I(x,y) & \quad x>y,
        \end{array}
    \right.
    $$
    for all $x,y \in [0,1]$.
    & The motivation is to modify a fuzzy implication function so that the new one satisfies \IP \cite{ZhangPei2017}. \Tstrut \\ \hline

    \textbf{\OP-modification} 

    Let $N$ be a fuzzy negation
    $$I^{(2)}(x,y) = \left\{
        \begin{array}{ll}
            1 & \quad x \leq y, \\
            I(x,y) \wedge (N(x) \vee y) & \quad x>y,
        \end{array}
    \right.
    $$
    for all $x,y \in [0,1]$.
    & The motivation is to modify a fuzzy implication function so that the new one satisfies \OP \cite{ZhangPei2017}. \Tstrut \\ \hline

    \textbf{Fuzzy implications based on semicopulas} 

    Let $B$ be a semicopula
    $$J_{I,B}(x,y) = I(x,B(x,y)),
    $$
    for all $x,y \in [0,1]$.
    & The motivation is to generalize probabilistic and survival implications and propose a new approach for constructing fuzzy implications \cite{Baczynski2017B,Zhao2019}.
    \Tstrut \\ \hline

    \textbf{$FNI$-implications} 

    Let $N$ a fuzzy negation and $F$ an aggregation function
    $$I_{FNI}(x,y) = F(N(x),I(x,y)),
    $$
    for all $x,y \in [0,1]$.
    & To generalize $I^N$ implications and to investigate how this method can be applied in such a way that the resulting fuzzy implication satisfies some additional desired properties not satisfied by the initial fuzzy implication \cite{Aguilo2018}.
    \Tstrut \\ \hline

    \textbf{Construction from ternary functions} 

    Let $F : [0,1]^3 \to [0,1]$ be a ternary linear polynomial function
    $$
    I_{F}(x,y) = F(x,y,I(x,y)),
    $$
    for all $x,y \in [0,1]$.
    & To present an easy construction method (based on linear functions of three variables) \cite{Massanet2018}.
    \Tstrut \\ \hline

    \textbf{Quadratic construction} 

    Let $F : [0,1]^3 \to [0,1]$ be a quadratic polynomial
    $$
    I_{F}(x,y) = F(x,y,I(x,y)),
    $$
    for all $x,y \in [0,1]$.
    & To present an easy construction method (based on quadratic functions of three variables) that preserves many of the most usual properties of fuzzy implication functions \cite{Kolesarova2019,Massanet2018}.
    \Tstrut \\ \hline

            $$  I_{I}^{0h}(x,y)  =
    \left\{ \begin{array}{ll}
    1 & x=0,\\
    0 &  x>0 \text{ and } y \leq e, \\
    I \left(ex, \frac{y-e}{1-e}\right) &  x>0 \text{ and } y >e,
    \end{array}
    \right.
     $$
                $$  I_{I}^{1h}(x,y)  =
    \left\{ \begin{array}{ll}
    1 & x=0,\\
    I\left(e+(1-e)x,\frac{y}{e}\right) &  x>0 \text{ and } y \leq e, \\
    1 &  x>0 \text{ and } y >e,
    \end{array}
    \right.
     $$
     
     Let $N$ be a fuzzy negation and $e \in (0,1)$

     \begin{itemize}
     \item If $N(x) \in [0,e]$ for all $e < x$ then
        $$  I_{I,N}^{1v}(x,y)  =
    \left\{ \begin{array}{ll}
    1 & x \leq e,\\
    I \left( N \left( \frac{N(x)}{e}\right),y\right) &  x>e,
    \end{array}
    \right.
     $$
        $$  I_{I,N}^{0h}(x,y)  =
    \left\{ \begin{array}{ll}
    0 & x>0, y\leq e,\\
    I \left(x , N \left( \frac{N(y)}{e}\right)\right) &  x>0, y>e.
    \end{array}
    \right.
     $$
         \item If $N(x) \in [e,1]$ for all $x<e$
        $$  I_{I,N}^{0v}(x,y)  =
    \left\{ \begin{array}{ll}
    1 & y=1,\\
    0 & x \geq e \text{ and } y<1,\\
    I \left(N \left(\frac{N(x)-e}{1-e}\right),y\right) &  x<e,
    \end{array}
    \right.
     $$
        $$  I_{I,N}^{1h}(x,y)  =
    \left\{ \begin{array}{ll}
    1 & y \geq e,\\
    I \left(x,N \left(\frac{N(y)-e}{1-e}\right)\right) & y<e.
    \end{array}
    \right.
     $$
     \end{itemize}
    & This two generation methods were introduced when studying properties that are preserved by the threshold horizontal \cite{Massanet2012A} and vertical methods of construction \cite{Massanet2013}.

    \Tstrut \\ \hline


   \textbf{} 
    Let $N$ be a fuzzy negation and $J$ a fuzzy implication function
    $$
    I(x,y) = J(1-N(x),y),
    $$
    for all $x,y \in [0,1]$.
    & To propose a new method of generating new fuzzy implications from a given one \cite{Souliotis2019}.
    \Tstrut \\ \hline

   \textbf{Max-operation}
   
    $$
    (I_1 \vee I_2)(x,y) = \max\{I_1(x,y),I_2(x,y)\},
    $$
    for all $x,y \in [0,1]$.
    & \cite[Theorem 6.1.1.]{Baczynski2008}
    \Tstrut \\ \hline

   \textbf{Min-operation}
   
    $$
    (I_1 \wedge I_2)(x,y) = \min\{I_1(x,y),I_2(x,y)\},
    $$
    for all $x,y \in [0,1]$.
    & \cite[Theorem 6.1.1.]{Baczynski2008}
    \Tstrut \\ \hline

   \textbf{Convex combination}
   
   Let $\lambda \in [0,1]$
    $$
    I_{I_1,I_2}^\lambda(x,y) = \lambda I_1(x,y) + (1-\lambda)I_2(x,y),
    $$
    for all $x,y \in [0,1]$.
    & \cite[Theorem 6.2.2.]{Baczynski2008}
    \Tstrut \\ \hline

   \textbf{$\circledast$-composition}  
    $$
    (I_1 \circledast I_2)(x,y) = I_1(x,I_2(x,y)), \quad x,y \in [0,1].
    $$
    & To give a rich algebraic structure on the set of all fuzzy implications $\mathbb{I}$. $(\mathbb{I},\circledast)$ is a non-idempotent monoid \cite{VemuriJayaram2012}.
    \Tstrut \\ \hline

   \textbf{$\triangledown$-composition}  
    $$
    (I_1 \triangledown I_2)(x,y) = I_1(I_2(y,x),I_2(x,y)), \quad x,y \in [0,1].
    $$
    & To give a rich algebraic structure on the set of all fuzzy implications $\mathbb{I}$. $(\mathbb{I},\triangledown)$ is a semigroup \cite{VemuriJayaram2012}.
    \Tstrut \\ \hline

   \textbf{$\odot_A$-composition}
   
   Let $A$ be an aggregation function
    $$
    (I_1 \odot_A I_2)(x,y) = A(I_1(x,y),I_2(x,y)), \quad x,y \in [0,1].
    $$
    & To give a rich algebraic structure on the set of all fuzzy implications $\mathbb{I}$. $(\mathbb{I},\odot_A)$ is a commutative monoid \cite{VemuriJayaram2012}.
    \Tstrut \\ \hline

   \textbf{$\diamond$-composition}
    $$
    (I_1 \diamond I_2)(x,y) = I_2(I_1(y,x),I_1(x,y)), \quad x,y \in [0,1].
    $$
    & To give a rich algebraic structure on the set of all fuzzy implications $\mathbb{I}$. $(\mathbb{I},\diamond)$ is a commutative monoid \cite{VemuriJayaram2012}.
    \Tstrut \\ \hline

   \textbf{}
    $$
    I_{I_1,I_2,I_3}(x,y) = I_1(I_2(y,x),I_3(x,y)), \quad x,y \in [0,1].
    $$
    & To generalize the following tautology of classical logic
    $$((b \to a) \to (a \to b)) \equiv a \to b$$
    \cite{Reiser2017}.\Tstrut \\ \hline

   \textbf{Rotation construction}
   
    Let $N$ be a strong fuzzy negation with $t$ its unique fixed point and $\mathscr{I}$ the linear transformation of $I$ into $[t,1]$, i.e., 
    $$\mathscr{I}_{[t,1]}(x,y) = t+(1-t)I \left(\frac{x-t}{1-t},\frac{y-t}{1-t}\right),
    $$
    and $D_{\mathscr{I}_{[t,1]}}$ the deresiduum, i.e., $D_{\mathscr{I}_{[t,1]}} = \inf \{z \in [t,1] \mid y \leq \mathscr{I}_{[t,1]}(x,z)\}$
    then

        $$
    I_{rot}(x,y)
    =
    \left\{ \begin{array}{ll}
    \mathscr{I}_{[t,1]}(N(y),N(x)) & x,y < t,\\
    1 &   x<t, y \geq t, \\
    \mathscr{I}_{[t,1]}(x,y) &   x,y \geq t, \\
    N(D_{\mathscr{I}_{[t,1]}}(x,N(y)) & x \geq t, y<t.
    \end{array}
    \right.
    $$
    \vspace{0.05cm}
    & To follow the same idea of rotation of t-norms but with fuzzy implication functions \cite{Su2018}.\Tstrut \\ \hline

   \textbf{Horizontal threshold generation method}

   Let $e \in (0,1)$
    $$
    I_{I_1 - I_2}^e =
    \left\{ \begin{array}{ll}
	1 & x=0,\\
	eI_1\left(x,\frac{y}{e}\right) &   x > 0 \text{ and } y \leq e, \\
    e + (1-e)I_2\left(x,\frac{y-e}{1-e}\right) &   x > 0 \text{ and } y > e.
\end{array}
\right.
    $$
    & To generalize the idea of the way of construction of $h$-implications in terms of an $f$ and a $g$-implication \cite{Massanet2012A}.
    \Tstrut \\ \hline

   \textbf{Vertical threshold generation method}

   Let $e \in (0,1)$
    $$
    I_{I_1 | I_2}^e =
    \left\{ \begin{array}{ll}
	e+(1-e) I_1 \left(\frac{x}{e},y\right) & x <e \text{ and } y < 1,\\
	e I_2 \left(\frac{x-e}{1-e},y\right) &   x \geq e \text{ and } y < 1, \\
    1 &   y=1.
\end{array}
\right.
    $$
    & To propose a new method to generate fuzzy implications from two given ones in the same spirit of the horizontal threshold generation method but now through an adequate scaling on the first variable of the given fuzzy implications \cite{Massanet2013}.
    \Tstrut \\ \hline

   \textbf{Aggregation of fuzzy implications}

   Let $A:[0,1]^n \to [0,1]$ be an aggregation function 
    $$
    I_A(x,y)
    =
    A(I_1(x,y),\dots,I_n(x,y)),
    $$
    for all $x,y \in [0,1]$.
    & To generalize the aggregation of fuzzy implication functions with an $n$-ary aggregation function $A$ and a family of fuzzy implications \cite{Reiser2013}.

    Some studies about the aggregation of specific families of fuzzy implication functions: \cite{Benitez2013,Benitez2014}.
    \Tstrut \\ \hline

   \textbf{Multiple fuzzy implications about $x$}

    $$
    J_{I_1,\dots,I_n}(x,y)
    =
    I_1(x,I_2(x,\dots,I_n(x,y))),
    $$
    for all $x,y \in [0,1]$.
    & To introduce a new way to generate a new fuzzy implication from given ones by means of multiple iteration about the first variable \cite{Li2017}.
    \Tstrut \\ \hline

   \textbf{Multiple fuzzy implications about $y$}
    $$
    G_{I_1,\dots,I_n}(x,y)
    =
    I_1(I_2(\dots I_n(x,y)\dots,y),y),
    $$
    for all $x,y \in [0,1]$.
    & To introduce a new way to generate a new fuzzy implication from given ones by means of multiple iteration about the second variable \cite{Li2017}.
    \Tstrut \\ \hline

   \textbf{Extended horizontal threshold generation method}

    Let $\{e_i\}_{i=1}^n$ be an increasing sequence in $(0,1)$
    $$
    I_E(x,y)
    =
    \left\{ \begin{array}{ll}
    1 & x=0,\\
    e_1I_1\left(x,\frac{y}{e_1}\right) &   x > 0, y \leq e_1, \\
    e_{i-1} + (e_{i}-e_{i-1})I_{i}\left(x,\frac{y-e_{i-1}}{e_{i}-e_{i-1}}\right) &   x > 0, e_{i-1} < y \leq e_{i}, \\
    e_n & x>0, y =e_n, \\
    e_n + (1-e_n)I_n \left(x,\frac{y-e_n}{1-e_n}\right) & x>0, y>e_n.
    \end{array}
    \right.
    $$
    for all $x,y \in [0,1]$.
    & To do a generalization of the threshold generation method by using a family of fuzzy implication functions in which the scaling method in the second variable is applied with different values \cite{Yi2017}.
    \Tstrut \\ \hline

   \textbf{Fuzzy implication functions based on $F$-chains}

    Let $F$ be an $n$-ary aggregation function and $c$ be an $F$-chain, i.e., $c : [0,1] \to [0,1]^n$ is an increasing mapping such that $c(0) = (0,\dots,0)$, $c(1)=(1,\dots,1)$ and $F(c(t))=t$ for all $t \in [0,1]$,
    $$
    I_{F,c}(x,y)
    =
    F(I(c_1(x),c_1(y)),\dots,I(c_n(x),c_n(y))),
    $$
    for all $x,y \in [0,1]$.
    & Similarly to the case of aggregation functions, the authors propose a new construction method of fuzzy implication functions based on $F$-chains \cite{Mesiar2019}.
    \Tstrut \\ \hline

   \textbf{$FI\varphi$-construction}

    Let $F$ be an aggregation function with $F(1,0)=0$ and $\varphi:[0,1] \to [0,1]$ a non-decreasing function with $\varphi(0)=0$ and $\varphi(1)=1$,
    $$
    I_{FI\varphi}(x,y)
    =
    F(I(x,y),\varphi(y)),
    $$
    for all $x,y \in [0,1]$.
    & To propose a new construction method that preserves many properties and to apply it to the construction of fuzzy subsethood measures \cite{Zhao2021}.
    \Tstrut \\ \hline




   \textbf{$R$-ordinal sums} 

   Expression in \cite[Theorem 5]{Mesiar2004}.
    & To describe the structure of residual implications generated by left-continuous t-norms as an ordinal sum of residual implications linked to the corresponding left-continuous t-subnorms that describe the structure of the corresponding left-continuous t-norm. In this sense, this is not a new family of fuzzy implication functions but is the first definition of the concept of ``ordinal sum of fuzzy implications'' \cite{Mesiar2004}. \Tstrut \\ \hline

   \textbf{Ordinal sum of fuzzy implications} 

   Expression in \cite[Definition 3.1]{Su2015b}.
    & To propose a definition of ordinal sum of fuzzy implication functions \cite{Su2015b}. \Tstrut \\ \hline
    
   \textbf{$N$-ordinal sums}

   Expression in \cite[Definition 4]{Massanet2016b}.
    & To generalize the structure of $(S,N)$-implications and to propose a new possible definition of ordinal sum of fuzzy implications considering $(S,N)$-implications derived from ordinal sums of t-conorms \cite{Massanet2016b}. \Tstrut \\ \hline

       \textbf{Ordinal sum I}
        \begin{center}
        \includegraphics[scale=0.5]{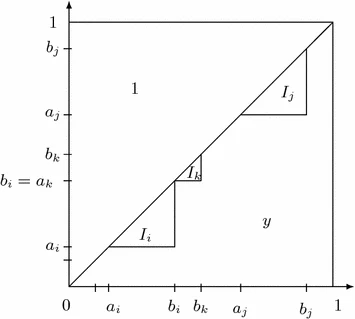}
        \end{center}
        Expression in  \cite[Definition 5]{Drygas2016}.
    & To generalize $R$-ordinal sums \cite{Drygas2016}. \Tstrut \\ \hline

       \textbf{Ordinal sum II}
        \begin{center}
        \includegraphics[scale=0.5]{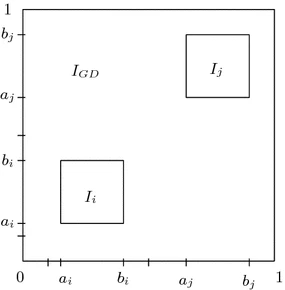}
        \end{center}
        Expression in  \cite[Definition 6]{Drygas2016}.
    & To propose an ordinal sum of fuzzy implication functions which is more similar to an ordinal sum of continuous, Archimedean t-norms. \cite{Drygas2016}. \Tstrut \\ \hline


           \textbf{Ordinal sum III}
        \begin{center}
        \includegraphics[scale=0.5]{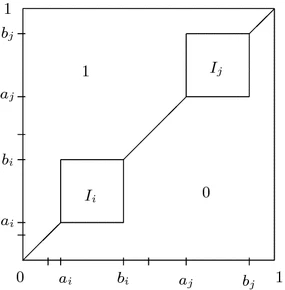}
        \end{center}
        Expression in  \cite[Definition 7]{Drygas2016}.
    & To define an ordinal sum of fuzzy implication functions in which any family of fuzzy implication functions can be used \cite{Drygas2016}. \Tstrut \\ \hline


           \textbf{Ordinal sum IV}
        \begin{center}
        \includegraphics[scale=0.5]{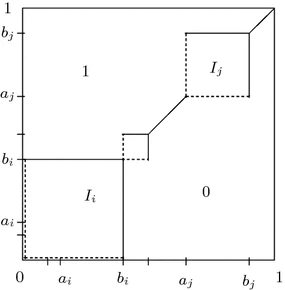}
        \end{center}
        Expression in  \cite[Definition 8]{Drygas2016}.
    & To define an ordinal sum of fuzzy implication functions in which the intervals are not separable \cite{Drygas2016}. \Tstrut \\ \hline


           \textbf{Ordinal sum V}
        \begin{center}
        \includegraphics[scale=0.5]{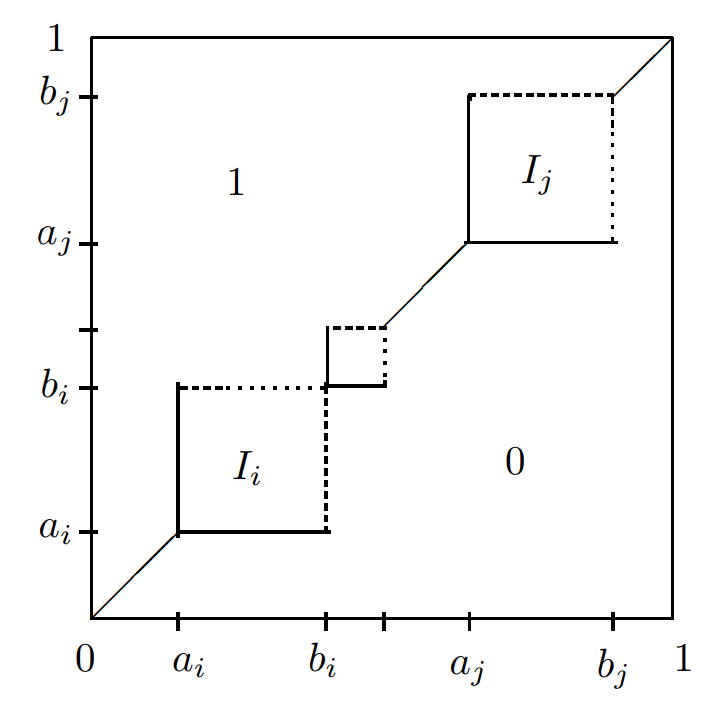}
        \end{center}
        Expression in  \cite[Definition 3.4]{Drygas2017}.
    & Another alternative to define an ordinal sum of fuzzy implication functions in which the intervals ar not separable \cite{Drygas2017}. \Tstrut \\ \hline


           \textbf{Ordinal sum VI}
        \begin{center}
        \includegraphics[scale=0.4]{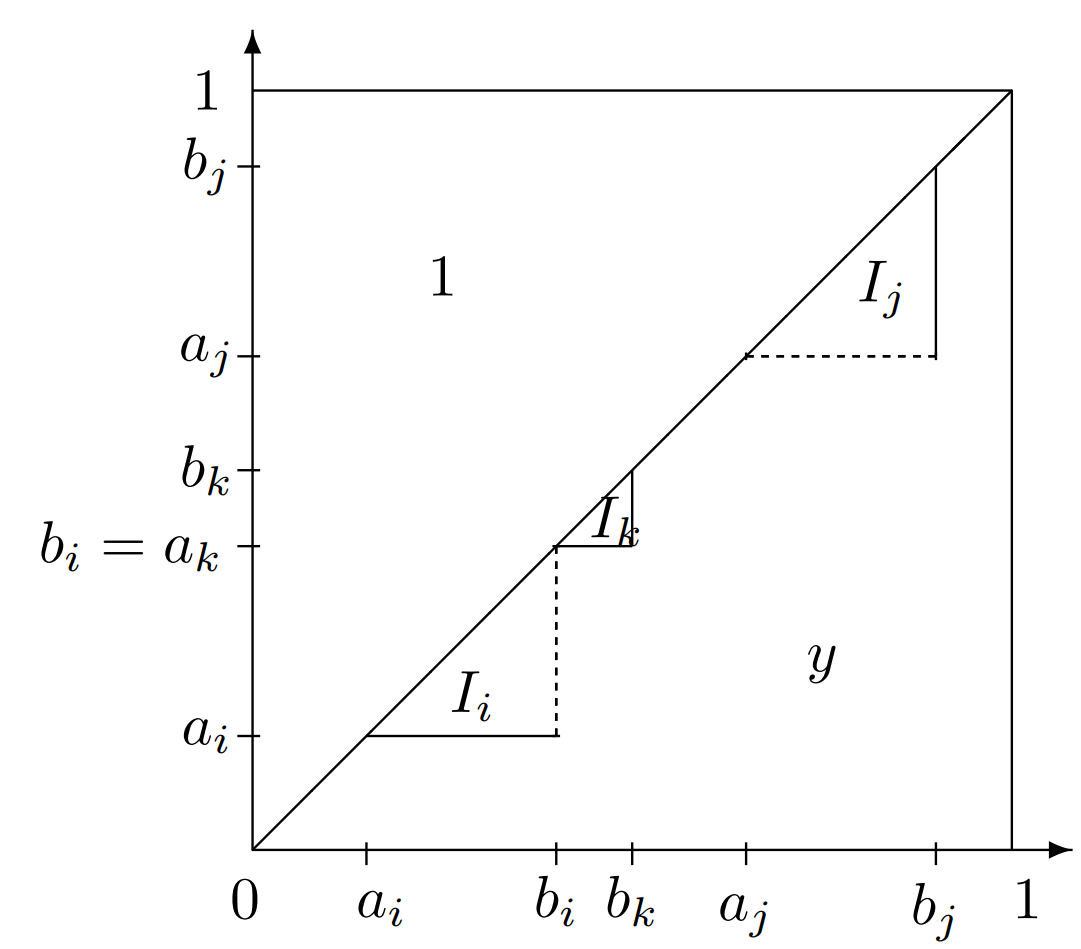}
        \end{center}
        Expression in  \cite[Definition 6]{Drygas2017b}.
    & To propose a generalization of Ordinal Sum I where they allow the intervals to be of different types: open, closed or half-open \cite{Drygas2017b}. \Tstrut \\ \hline


           \textbf{Ordinal sum VII}
        \begin{center}
        \includegraphics[scale=0.6]{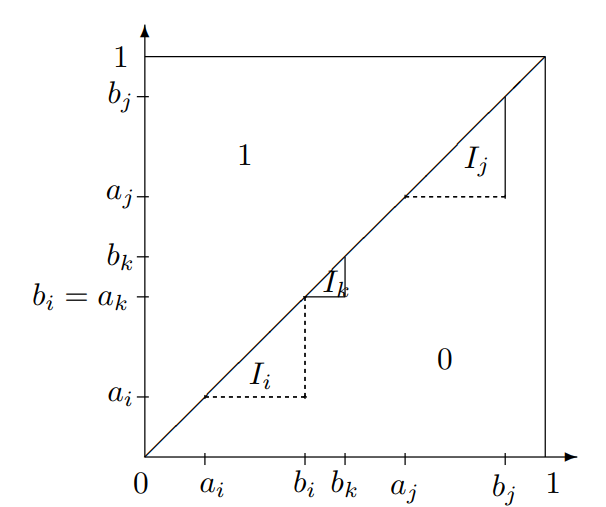}
        \end{center}
        Expression in  \cite[Definition 7]{Drygas2017b}.
    & To propose an alternative to Ordinal Sum VI changing the values of the generated implications outside the fixed intervals \cite{Drygas2017b}. \Tstrut \\ \hline


           \textbf{Ordinal sum VIII}
        \begin{center}
        \includegraphics[scale=0.6]{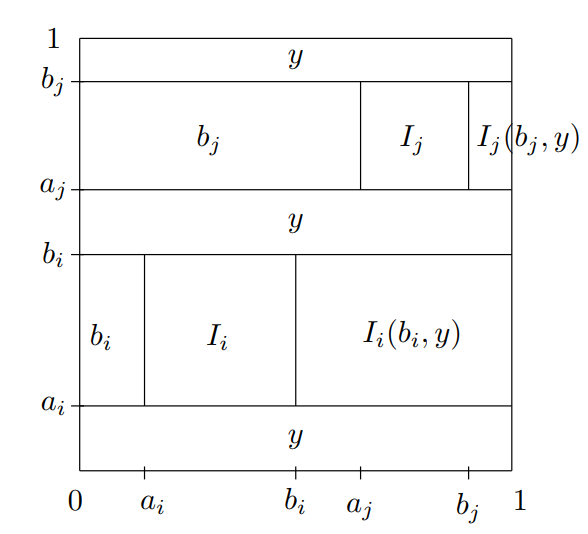}
        \end{center}
        Expression in  \cite[Definition 18]{Baczynski2017}.
    & To introduce new ways of constructing ordinal sums of fuzzy implication functions based on a construction of ordinal sums of overlap functions \cite{Baczynski2017}. \Tstrut \\ \hline


           \textbf{Ordinal sum IX}
        \begin{center}
        \includegraphics[scale=0.6]{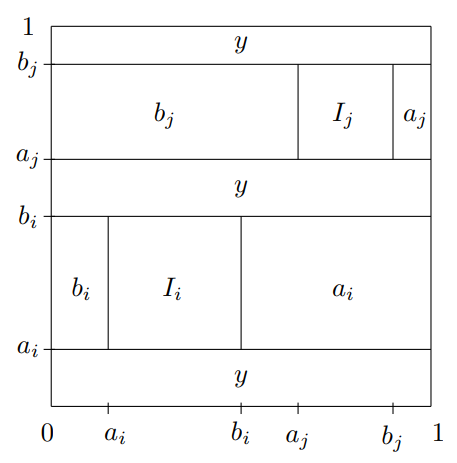}
        \end{center}
        Expression in  \cite[Definition 15]{Baczynski2017}.
    & To introduce new ways of constructing ordinal sums of fuzzy implication functions based on a construction of ordinal sums of overlap functions \cite{Baczynski2017}. \Tstrut \\ \hline


           \textbf{$F$-chain based ordinal sums of fuzzy implication functions}
        Expression in  \cite[Theorem 4.1]{Mesiar2019}.
    & To define a new ordinal sum of fuzzy implication functions based on $F$-chains \cite{Mesiar2019}. \Tstrut \\ \hline


           \textbf{Left ordinal sum of fuzzy implications}
           
        Expression in  \cite[Definition 5.1]{Lu2020}.
    & Previous fuzzy implication functions defined as an ordinal sum of other fuzzy implication functions have as natural negation the bottom negation. The authors propose a new definition of ordinal sum of fuzzy implication functions whose natural negation corresponds to an ordinal sum of the corresponding natural negations of the family of fuzzy implication functions involved \cite{Lu2020}. \Tstrut \\ \hline


           \textbf{Implication complementing and reconstructing}
           
    & To present two construction ways to study the general forms of ordinal sums of fuzzy implications with the intent of unifying the ordinal sums existing in the literature \cite{Zhou2020}. \Tstrut \\ \hline
    
\caption{Families of fuzzy implication functions generated by other fuzzy implication functions.}\label{table:construction_methods}
\end{longtable}




\section{Conclusions and future work}\label{sec:conclusions}

In this survey, we have provided an overview of additional properties of fuzzy implication functions that have been considered in the literature, along with a comprehensive compilation of several families, classified by their construction method. Our main objective was to offer a structured and accessible document that facilitates both theoretical and practical research on these operators. By collecting and classifying a wide range of families and properties, it is now easier to have a general picture of what is done in the literature and to address new problems. However, this document is not enough to quantify the number of different existing families, because these families can present intersection or even coincide \cite{Massanet2017B}. That is why it is of the utmost importance to study the additional properties that the operators of a certain family satisfy and to provide an axiomatic characterization of the new operators in the literature in order to find its possible relation with respect to those already known \cite{Massanet2024}. Further, due to the extensive literature on the topic, there may be other families that we have missed.

In future work, we want to study and include in this document more information about the gathered families, like the main properties studied for each family, their characterizations, and intersections. Also, for the construction methods we can study which additional properties are preserved. With this information, it will be easier to compare families, identify the strengths of each one compared to the others, and disclose new problems in the field. Also, it would be interesting to keep a record of which families have already been considered for which particular applications.

\section*{Acknowledgment}

Raquel Fernandez-Peralta is funded by the EU NextGenerationEU through the Recovery and Resilience Plan for Slovakia under the project No. 09I03-03-V04- 00557.
\bibliographystyle{unsrt}  
\bibliography{arxiv}  

\end{document}